\documentclass{article}
\PassOptionsToPackage{numbers, compress}{natbib}

\usepackage[final]{neurips_2022}

\usepackage[utf8]{inputenc} 
\usepackage[T1]{fontenc}    
\usepackage{hyperref}       
\usepackage{url}            
\usepackage{booktabs}       
\usepackage{amsfonts}       
\usepackage{nicefrac}       
\usepackage{microtype}      

\usepackage{microtype}
\usepackage{graphicx}
\usepackage{subfigure}
\usepackage{booktabs}
\usepackage{enumitem}
\usepackage[normalem]{ulem}
\useunder{\uline}{\ul}{}

\usepackage{sidecap} 
\sidecaptionvpos{figure}{t}
\sidecaptionvpos{table}{t}

\usepackage{wrapfig}

\usepackage[table,xcdraw]{xcolor}
\usepackage{xcolor}
\usepackage{amsmath} 
\usepackage{pifont} 
\usepackage{sidecap} 
\usepackage{varwidth} 
\usepackage{amsfonts}
\usepackage{booktabs}
\newcommand{\walon}[1]{\textcolor{black}{#1}}
\newcommand{\new}[1]{\textcolor{black}{#1}}

\newcommand{\nips}[1]{\textcolor{black}{#1}}
\newcommand{\modify}[1]{\textcolor{black}{#1}}
\newcommand{\appendixNew}[1]{\textcolor{black}{#1}}
\usepackage{comment}
\usepackage{multirow} 

\usepackage[ruled,vlined]{algorithm2e}

\usepackage{color}

\newcommand{\rebuttal}[1]{\textcolor{black}{#1}}
\newcommand{\CR}[1]{\textcolor{black}{#1}} 

\title{Make an Omelette with Breaking Eggs:\\ Zero-Shot Learning for Novel Attribute Synthesis}

\author{%
  Yu-Hsuan Li$^\ast$ \\
  National Chiao Tung University\\
  \texttt{evali890227@gmail.com} \\
  \And
  Tzu-Yin Chao$^\ast$ \\
  National Chiao Tung University\\
  \texttt{chaotzuyin@nctu.edu.tw} \\
  \AND
  Ching-Chun Huang \\
  National Chiao Tung University\\
  \texttt{chingchun@cs.nctu.edu.tw} \\
  \And
  Pin-Yu Chen \\
  IBM Research \\
  \texttt{pin-yu.chen@ibm.com} \\
  \And
  Wei-Chen Chiu \\
  National Chiao Tung University \\
  \texttt{walon@cs.nctu.edu.tw} \\
}

\newcommand{\nocontentsline}[3]{}
\newcommand{\tocless}[2]{\bgroup\let\addcontentsline=\nocontentsline#1{#2}\egroup}

\begin{document}
\maketitle

\tocless
\begin{abstract}

\new{
    Most of the existing algorithms for zero-shot classification problems typically rely on the attribute-based semantic relations among categories to realize the classification of novel categories without observing any of their instances. 
    However, training the zero-shot classification models still requires attribute labeling for each class (or even instance) in the training dataset, which is also expensive. To this end, in this paper, we bring up a new problem scenario:
    \walon{``\textit{\modify{Can we} derive zero-shot learning for novel attribute detectors/classifiers and use them to automatically annotate the dataset for labeling efficiency?}''.}
    \walon{Basically, given only a small set of detectors that are learned to recognize some manually annotated attributes (i.e., the seen attributes), we aim to synthesize the detectors of novel attributes in a zero-shot learning manner.} 
    Our proposed method, \textbf{Z}ero-\textbf{S}hot \textbf{L}earning for \textbf{A}ttributes (ZSLA), which is the first of its kind to the best of our knowledge, tackles this new research problem by applying the set operations to first decompose the seen attributes into their basic attributes and then recombine these basic attributes into the novel ones.
    Extensive experiments are conducted to verify the capacity of our synthesized detectors for accurately capturing the semantics of the novel attributes and show their superior performance in terms of detection and localization compared to other baseline approaches. 
    \modify{Moreover, we demonstrate the application of automatic annotation using our synthesized detectors on Caltech-UCSD Birds-200-2011 dataset. Various generalized zero-shot classification algorithms trained upon the dataset re-annotated by ZSLA shows comparable performance with those trained with the manual ground-truth annotations.}} Please refer to our project page for source code: \url{https://yuhsuanli.github.io/ZSLA/}

\end{abstract}
\tocless\section{Introduction}

\begin{figure*}[t] 
\centering 
\includegraphics[width=1\textwidth]{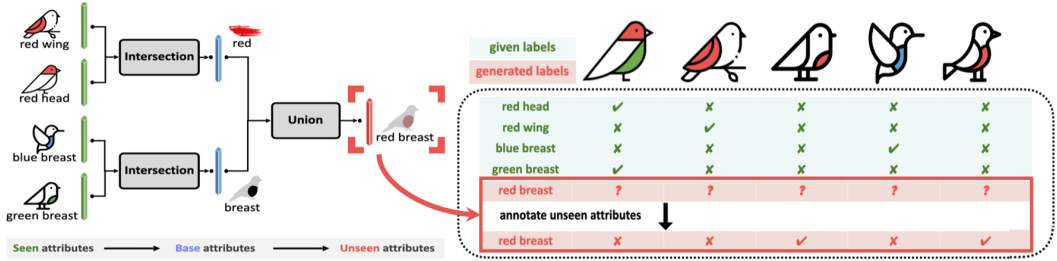} 
\vspace{-1em}
\caption{Given a set of trained/seen attribute detectors (e.g. ``red wing", ``red head", ``blue breast", and ``green breast"), our ZSLA can synthesize a novel detector for the unseen attribute (e.g. ``red breast") by the following process: 
(1) applying the intersection operation on the subsets $\{$``red wing", ``red head"$\}$ and $\{$``blue breast", ``green breast"$\}$ respectively to extract the common semantics of each subsets, i.e. ``red" and ``breast", as the \emph{\textbf{base attributes}}; (2) combining the base attributes via the union operation to realize the novel/unseen attribute detector, i.e. ``red breast". The novel attribute detectors can later be applied to annotate the dataset.
} 
\label{Fig.teaser}       
\vspace{-0.5em}
\vspace{-0.5cm}
\end{figure*}



\par \new{Zero-shot learning (ZSL) algorithms for classification aim to recognize novel categories without observing any of their instances during model training; thus, the cost of collecting training samples for the novel categories can be eliminated. Typically, the core challenge behind zero-shot classification lies in associating novel categories with the seen ones during training. Various existing approaches leverage different auxiliary semantic information to construct such associations across categories, thus being able to generalize the learned models for classifying novel categories~\cite{DAP_IAP, ALE, ESZSL, SAE, LAGO, AGZSL} or synthesize the training samples for each novel category~\cite{f-WGAN, f-VAEGAN-D2, CADAVAE, EPGN, tfVAEGAN}. Among different types of auxiliary semantic information adopted for ZSL, defining a group of attributes shared among categories becomes one of the most popular choices, where each category is described by multiple attributes (i.e., multi-labeled by the attributes), and the attribute-based representations are discriminative across categories.
However, it comes with the expensive cost of manually annotating the samples in the dataset their attribute labels at a much granular level.
For example, CUB dataset~\cite{CUB}, one of the most widely-used benchmarks for learning zero-shot classification, is built by spending a great deal of time and effort to label 312 attributes for 11788 images.}

\par \new{As motivated by the issue of annotation efficiency on attribute labels, this paper aims to \emph{\textbf{develop ZSL on known attributes to annotate novel attributes for a dataset automatically 
}}.
That is, analogous to the zero-shot classification scenario, we now advance to annotate novel attributes for a dataset via utilizing the knowledge from a few types of seen/given manual attributes, as illustrated in Figure \ref{Fig.teaser}. Specifically, we take the well-known CUB dataset~\cite{CUB} as our main test-bed and have a deep investigation on its attributes. We discover that, many attributes in CUB dataset (e.g. ``red head" or ``blue belly") follow the form of combinations over \emph{\textbf{base attributes}} (e.g. ``red", ``blue", ``head" and ``belly" respectively).
Building upon such observation, given a defined set of attributes in the form of the ones used in the CUB dataset and labels of a few \emph{\textbf{seen attributes}} (where the number is far less than that of overall defined attributes), we propose \textbf{Z}ero-\textbf{S}hot \textbf{L}earning for \textbf{A}ttributes (ZSLA), a method of training the \emph{\textbf{seen attribute detectors}} and then tackle the ZSL problem to synthesize unseen attribute detectors via a \emph{\textbf{decompose-and-reassemble}} manner. In detail, the seen attribute detectors are firstly decomposed into base attribute representations, in which they are further reassembled with novel combinations into novel attribute detectors, as illustrated in Figure~\ref{Fig.teaser}. Here, both the decomposition and reassembly steps are achieved via set operations (i.e., the interaction and union operators, respectively). Together with the seen ones, the novel attribute detectors can be utilized to annotate the attribute labels for the dataset automatically.
}


\par \new{
To demonstrate the efficiency of ZSLA, we synthesize 207 novel attribute detectors by leveraging only 32 seen ones from the CUB dataset. 
These novel attribute detectors are shown to be effective in capturing their corresponding semantic information and benefit both the attribute detection and localization for the samples in CUB dataset.
Besides, we also synthesize $\alpha$-CLEVR dataset by \cite{clevr} for conducting the controlled experiments to further discuss the influence of noisy seen attribute labels. The results show that ZSLA can provide more robust annotations than the other baseline methods under the noisy scenario. Below, we highlight the contributions of this paper:
}
%


\begin{itemize}[leftmargin=*]
\vspace{-0.5em}
\item To the best of our knowledge, we are the first to propose ZSL for attributes to automatically annotate attribute labels for the zero-shot classification datasets.
\vspace{-0.5em}
\item We propose a novel decompose-and-reassemble approach to single out the base attribute representations \modify{by}  applying intersection on the seen attribute detectors and synthesize the unseen ones by having the union operation over the base attributes representations.
\vspace{-0.5em}
\item \modify{In Section \ref{sec:reannotate}}, we show on the CUB dataset that, given only 32 attributes with manual annotations, ZSLA can synthesize novel attribute detectors to provide high-quality annotations for the dataset. By using the auto-annotated attributes, generalized zero-shot classification algorithms can also achieve comparable or even better performance than that using 312 manually-annotated attributes.
\end{itemize}

\tocless\section{Related Works}
\walon{Zero-shot learning (ZSL) was originally proposed to tackle the specific classification problem, where the model is expected to be capable of classifying the samples belonging to the novel categories which are not seen previously during training. The problem setup has been extended to other applications such as detection~\cite{bansal2018zero, rahman2018zero, demirel2018zero} and segmentation~\cite{bucher2019zero, zheng2021zero}. Here we provide a brief review of the works of zero-shot classification~\cite{DAP_IAP, ESZSL, ALE, SJE, SAE, xu2020attribute,f-WGAN, f-VAEGAN-D2, CADAVAE, EPGN, tfVAEGAN}. Without loss of generality, the ZSL approaches rely on using the auxiliary information (such as attributes, word embeddings, or text descriptions) as the basis for describing the categories and building the semantic relation among seen and unseen categories; and the existing methods can be roughly categorized into two groups: the embedding-based methods~\cite{ALE, SJE, SAE, CADAVAE, xu2020attribute} and generative methods~\cite{f-WGAN, f-VAEGAN-D2, CADAVAE, EPGN, tfVAEGAN}. The embedding-based methods basically aim to learn a latent space that connects between the feature representations of training samples and the embeddings of their corresponding auxiliary information (e.g., the visual features and the embeddings of attribute labels for the training images in the CUB dataset), such that the test samples can be classified as the novel categories once their feature representations are close to the embeddings of novel categories (which are defined upon auxiliary information without requiring any additional training samples). The generative methods instead utilize the deep-generative models (e.g., generative adversarial networks~\cite{goodfellow2014generative}, variational autoencoder~\cite{kingma2013auto}, or their hybrids/variants) for learning to synthesize the samples or features of the unseen categories based on their auxiliary semantic information. Though saving the effort of collecting the training samples to recognize novel categories via ZSL techniques, manually annotating the auxiliary semantic information for the samples in the zero-shot training dataset is still quite expensive and time-consuming. \nips{In turn, our proposed task of zero-shot learning for novel attributes helps to reduce such costs for the scenario of zero-shot classification where the auxiliary information is defined on attributes.}}

\CR{Aside from the typical zero-shot classification problem, \textit{compositional zero-shot learning} (CZSL) and \textit{blind source separation}~(BSS) are tasks that our work is conceptually related to. CZSL~\cite{misra2017red, nagarajan2018attributes, atzmon2020causal, mancini2021open, huynh2020compositional, naeem2021learning}, also as known as \textit{state-object compositionality} problem, aims to recognize the novel compositions (e.g. ``ripe tomato'') given the seen visual primitives of states 
 (e.g. ``ripe'', ``rotten'') and objects (e.g. ``apple'', ``tomato'') in the training dataset.}
Our proposed problem scenario ZSLA is distinct from CZSL under several perspectives: (1) An image in our problem scenario would have multiple attributes while there usually exists only a single state-object composition for CZSL; (2) Our synthesized attribute detectors are able to provide labels of novel attributes (i.e. these novel attributes do not have any manually labeled samples in the training set) for the images thus leading to more detailed descriptions for all the categories, while CZSL typically aims to increase the number of categories (i.e. each novel composition is treated as a new fine-grained class). 
\CR{We provide the extensive discussions and experiments with respect to CZSL in Appendix \ref{appendix:CZSL}.}

\CR{On the other hand, BSS\cite{comon1994independent, wang2008time, yilmaz2004blind, delfosse1995adaptive} aims to separate and distinguish the source of several signals from their mixture. }
\CR{Though our ZSLA framework composed of a decompose-and-reassemble procedure seems to be similar to blind source separation at first glance, there exists a significant distinction that differentiates our ZSLA from BSS in signal processing: Our intersection operation to perform decomposition on seen attributes is non-blind, in which it works by the guidance of logical and semantic constraints (i.e. two input attributes should have a common ground in one of the base attributes but not both); by contrast, there is no such constraint in the blind source separation (which instead typically adopts independent assumption or mutual information in its modeling).}

\tocless\section{ZSLA: Proposed Method}

\walon{Given a zero-shot classification dataset $\{\mathbf{X}, \mathbf{Y}, \mathbf{A}^s\}$, each image $x \in \mathbf{X}$ has its class label $y \in \mathbf{Y}$ and the multi-attribute labels $\phi^s(x)$, where $\phi^s(x)$ is a binary vector with its each element denoting if $x$ has a certain attribute $a \in \mathbf{A}^s$. ZSLA starts with using $\{\mathbf{X}, \mathbf{A}^s\}$ to train the detectors $M^s$ for all the attributes in $\mathbf{A}^s$, which are treated as seen attributes, then it adopts the seen attribute detectors $M^s$ to synthesize the detectors $M^u$ for the unseen attributes $\mathbf{A}^u$ via a decompose-and-reassemble procedure, where $\mathbf{A}^s \cap \mathbf{A}^u = \emptyset$.} \modify{For ease of understanding,} we use the most popular zero-shot classification dataset, CUB~\cite{CUB}, to illustrate how these steps are realized as follows.


\subsection{Training Seen Attribute Detectors}
\label{sec: stage_1}
\walon{
Our attribute detectors are built on top of the image feature space produced by the image feature extractor $f$. Given an input image $x$ and its feature map $f(x) \in \mathbb{R}^{W\times H\times C}$ where each $C$-dimensional feature vector at position $(i, j)$ of $f(x)$, denoted as $f(x)[i, j]$, is the feature representation of the corresponding image patch on $x$, the attribute detectors $M^s \in \mathbb{R}^{C \times N^s}$ (in which $N^s$ denotes the number of attributes in $\mathbf{A}^s$) aim to give high response on the image patches containing the visual appearance related to the attributes in $\mathbf{A}^s$. Specifically, each column in $M^s$ is acting as the embedding of a certain attribute. We use $m^s_k$ to indicate the $k$-th column of $M^s$. The response of the corresponding $k$-th attribute in $\mathbf{A}^s$ 
with respect to the patch-wise feature vector $f(x)[i, j]$ is calculated by 
a specific form of their cosine similarity $\texttt{cos}(\left|m^s_k\right|, f(x)[i, j])$, where $\left|m^s_k\right|$ denotes applying element-wise absolute-value operator on $m^s_k$. We have $\left|m^s_k\right|$ in our cosine similarity computation due to the reason that: Each dimension along channels of $f(x)$ is considered to capture a specific visual pattern. Our $\left|m^s_k\right|$ hence acts as to apply the weighted combination over these various visual patterns for representing the characteristics of the $k$-th attribute in $\mathbf{A}^s$, and the absolute-value operator over $m^s_k$ is to ensure the combination weights are non-negative.}

\begin{figure*}[ht] 
\centering 
\includegraphics[width=0.95\textwidth]{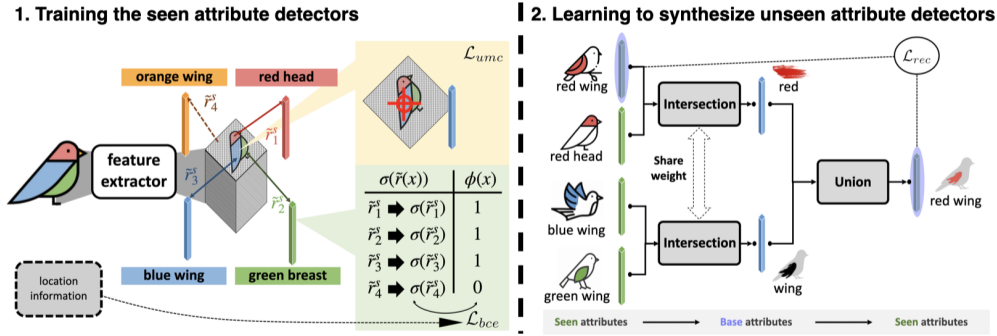} 
\caption{\walon{Overview of Our ZSLA}. \textbf{(1)}~Training the seen attribute detectors: \walon{Seen attribute detectors, defined as the embeddings for each seen attribute, are built on top of the image features and their training is guided by two objectives: $\mathcal{L}_{bce}$ and $\mathcal{L}_{umc}$, where the former drives the trained detectors to perform binary classification for attributes on image patches (cf. Eq.~\ref{eq:seen_ce_no_location}) while the latter enforces the uni-modal constraint on the response map $\mathcal{R}^s(x)$ of patch-wise image features with respect to each attribute, in order to make it compact and concentrated (cf. Eq.~\ref{eq:umc}).} \textbf{(2)}~Learning to synthesize novel/unseen attribute detectors via a decompose-and-reassemble procedure: 
\walon{Given the trained detectors of seen attributes, the intersection operation is firstly applied on them to extract base attributes, and then these base attributes are further combined by union operation to synthesize the novel/unseen attributes. The training of these operations is driven by the reconstruction loss $\mathcal{L}_{rec}$ (cf. Eq.~\ref{eq:rec}) once the synthesized attribute coincides with any of the seen ones. 
}
}
\label{Fig.train_process} 
\end{figure*}

\new{
We denote $\mathcal{R}^s(x) \in \mathbb{R}^{W\times H \times N^s}$ as the response map which has included the cosine similarities of all the seen attributes $\mathbf{A}^s$ at each position on $f(x)$. Note that, as our feature extractor $f$ adopts the ReLU activation function in its last layer (similar to most image feature extractors based on the convolutional networks), the values in $f(x)$ become non-negative. Furthermore, as both $\left|m^s_k\right|$ and $f(x)[i, j]$ are non-negative vectors, all entries of $\mathcal{R}^s$ results to be within the range $[0, 1]$. Following the popular tricks for ZSL and deep learning pointed out in~\cite{skorokhodov2020class}, where adopting scaled cosine similarity in logits computation is important to achieve better model training, we use the computation below to calibrate the value of elements in $\mathcal{R}^s(x)$: 
\begin{equation}
    \tilde{R}^s(x) = \gamma^2 \cdot (2\cdot R^s(x) -1 )
    \label{eq:scaled_cos_sim}
\end{equation}
where the calculation within brackets shifts and expands the values in $R^s(x)$ towards $[-1, 1]$ to match the typical value range of cosine similarity, and the hyperparameter $\gamma$ is set to 5 as suggested by~\cite{skorokhodov2020class}.
Then, we perform the max-pooling operation on $\tilde{R}^s(x)$ and obtain the image-wise attribute response $\tilde{r}^s(x) \in \mathbb{R}^{N^s}$.
Such logits over attributes thus are able to drive the model training (i.e. optimization over $M^s$ and $f$) via the error between the attribute detection results and the ground-truth attribute labels $\phi^s(x)$. The objective function $\mathcal{L}_{bce}$ to evaluate the error between the logits of attribute detection result $\tilde{r}^s(x)$ and the ground-truth attribute labels $\phi^s(x)$ is defined via the binary cross-entropy:
\vspace{-0.5em}
\begin{equation}
    \begin{aligned}
    \mathcal{L}_{bce} = -\sum \limits_{k}^{N^s} &\phi^s_k(x) \cdot \log(\sigma(\tilde{r}^s_k(x))) + (1-\phi^s_k(x)) \cdot \log(1-\sigma(\tilde{r}^s_k(x)))
    \label{eq:seen_ce_no_location}
    \end{aligned}
\end{equation}
where $\phi^s_k(x)$ and $\tilde{r}^s_k(x)$ denote the $k$-th elements in $\phi^s(x)$ and $\tilde{r}^s(x)$ respectively, and $\sigma$ is the sigmoid function.
}

In addition to the $\mathcal{L}_{bce}$ loss, we introduce another objective function $\mathcal{L}_{umc}$ to place \modify{the \emph{\textbf{uni-modal constraint}} ~\cite{zheng2017learning}} on the response map $\tilde{R}^s(x)$, which 
encourages the response map for a certain attribute (e.g. $\tilde{R}_k^s(x)$, the $k$-th channel of $\tilde{R}^s_k(x)$) to be uni-modal and concentrated. 
\modify{In other words, we regularize a detector to focus on a single location or a small region in the image $x$. We provide a further discussion of relaxing this assumption in \CR{Appendix~\ref{appendix:UMC}}.
}
\begin{equation}
    \begin{aligned}
    \mathcal{L}_{umc} = \sum^{N^s}_k\sum_{(i,j)} \sigma(\tilde{R}^s_k(x)[i, j]) \cdot (\left\| i-\breve{i}_k \right\|^2 + \left\| j-\breve{j}_k \right\|^2),
    \end{aligned}
    \label{eq:umc}
\end{equation}
   where $\breve{i}_k, \breve{j}_k = \mathop{\arg\max}_{i, j} \tilde{R}^s_k(x)[i,j]$ and $\|\cdot\|$ denotes the Euclidean norm.

\new{
The overall objective to train the feature extractor $f$ and the seen attribute detectors $M^s$ is illustrated in the left portion of Figure~\ref{Fig.train_process} and summarized as:
$\mathcal{L}_{bce}+\lambda\mathcal{L}_{umc}$,
where the hyperparameter $\lambda$ controls the balance between losses and is set to $0.2$ in our experiments.}

Moreover, we are aware that in CUB dataset the additional annotations indicating the ground-truth locations for the attributes which an image $x$ has are also available (e.g. we know where the attribute ``brown wing'' appears on an image of ``gadwall''). Hence, in addition to max-pooling the response map $\mathcal{R}^s(x)$ to obtain the image-wise response $r^s(x)$ for attributes, we experiment another way to obtain $r^s(x)$: (1) If $\phi^s_k(x)$ is true, the $k$-th element in $r^s(x)$, i.e. $r^s_k(x)$, is assigned by $\mathcal{R}^s(x)[i,j]$ where the centre of the ground-truth location for the $k$-th attribute in $\mathbf{A}^s$ is located on the patch related to the position $(i, j)$ of $\mathcal{R}^s$; (2) If $\phi^s_k(x)$ is false, $r^s_k(x)$ is assigned by having the average pooling over the $k$-th channel of $\mathcal{R}^s(x)$. \CR{Appendix \ref{appendix:ablation study} provides} the analysis for the impact of using such additional annotations of attribute location on the performance of ZSLA.

\subsection{Decompose-and-Reassemble for Synthesizing Novel Attribute Detectors}
\label{sec:DandR}
After obtaining the seen attribute detectors $M^s$, we now aim to perform the decompose-and-reassemble procedure (as shown in the right-half of Figure~\ref{Fig.train_process}) for generating the detectors $M^u \in \mathbb{R}^{C\times N^u}$ of the novel attributes $\mathbf{A}^u$ (where $N^u$ is the number of attributes in $\mathbf{A}^u$) by leveraging $M^s$.


\walon{First, we observe that most of the attributes in CUB dataset (the most popular zero-shot classification dataset and also our test-bed in this work) follow the form of ``\textit{adjective} + \textit{object part}'', for instance: ``black eye'', ``brown forehead'', ``red upper-tail'', or ``buff breast''. Starting from such observation, we define two disjoint sets of \emph{\textbf{base attributes}}, $\mathbf{B}^c$ and  $\mathbf{B}^p$, representing the \textit{adjectives} and \textit{object parts} used in the seen attributes,  respectively (e.g. ``blue'', ``yellow'', ``solid'', and ``perching-like'' for $\mathbf{B}^c$;  ``leg'', ``beak'', ``belly'', and ``throat'' for $\mathbf{B}^p$). Please note that the concepts behind adjectives $\mathbf{B}^c$ in CUB dataset include not only color but also texture, shape, and others. Formally, given an attribute $a$, we use $\beta^c(a)$ and $\beta^p(a)$ to denote its corresponding base attributes on the adjective and object part,  respectively (i.e. $\beta^c(a) \in \mathbf{B}^c$ and $\beta^p(a) \in \mathbf{B}^p$), where $\beta^c(\cdot)$ and $\beta^p(\cdot)$ are functions to indicate the base attributes in $\mathbf{B}^c$ and $\mathbf{B}^p$ for an attribute $a$, respectively.}

\walon{Now, given two seen attributes $a_k$ and $a_l \in \mathbf{A}^s$ in which $a_k = \{\beta^c(a_k), \beta^p(a_k)\}$ and $a_l = \{\beta^c(a_l), \beta^p(a_l)\}$, if $a_k$ and $a_l$ have common ground in either the base attribute of adjectives (i.e. $\beta^c(a_k) = \beta^c(a_l) \in \mathbf{B}^c$) or the one of object parts (i.e. $\beta^p(a_k) = \beta^p(a_l) \in \mathbf{B}^p$) but not both, then we can use the \emph{\textbf{intersection operation}} $\mathbb{I}$ to extract such common base attribute from $a_k$ and $a_l$:
\begin{equation}
\label{eq:intersection}
\begin{aligned}
    \mathbb{I}(&a_k, a_l) = &\begin{cases}
\beta^c(a_k)  & \text{ if } \beta^c(a_k) = \beta^c(a_l),~\beta^p(a_k) \neq \beta^p(a_l)\\ 
\beta^p(a_k)& \text{ if } \beta^c(a_k) \neq \beta^c(a_l),~\beta^p(a_k) = \beta^p(a_l)
\end{cases}
\end{aligned}
\end{equation}
For instance, $\mathbb{I}$ is able to extract the base attribute ``red'' from the seen attributes ``red wing'' and ``red breast''; or the base attribute ``tail'' from the seen attributes ``buff tail'' and ``black tail''.
}

\walon{Once we obtain the base attributes via intersection over seen attributes, we further adopt the \emph{\textbf{union operation}} $\mathbb{U}$ to create novel attributes. Given two pairs of seen attributes $\{a_k, a_l\}$ and $\{a_{{k}'}, a_{{l}'}\}$ in which $\beta^c(a_k) =  \mathbb{I}(a_k, a_l)$ and $\beta^p(a_{{k}'}) =  \mathbb{I}(a_{{k}'}, a_{{l}'})$, i.e. $\{a_k, a_l\}$ share the same base attribute of adjective while $\{a_{{k}'}, a_{{l}'}\}$ share the same base attribute of object part, a novel attribute $\tilde{a}$ can be synthesized by combining $\beta^c(a_k)$ and $\beta^p(a_{{k}'})$, i.e. $\tilde{a} = \mathbb{U}(\beta^c(a_k), \beta^p(a_{{k}'}))$. In particular, if such combination of base attributes has been seen in $\mathbf{A}^s$, i.e. there exists an attribute $a \in \mathbf{A}^s$ where $\beta^c(a) = \beta^c(\tilde{a})$ and $\beta^p(a) = \beta^p(\tilde{a})$, we say the seen attribute $a$ is \emph{\textbf{reconstructed}} by $\tilde{a}$. Otherwise, if none of the seen attributes has the identical combination as our synthesized $\tilde{a}$, we denote $\tilde{a}$ a \emph{\textbf{novel attribute}} and $\tilde{a} \in \mathbf{A}^u$. In summary, extracting base attributes from seen attributes via intersection, followed by combining the base attributes into novel attributes via union, holistically forms our \emph{\textbf{decompose-and-reassemble}} procedure to synthesize the novel attributes.}

In practice, our implementation of the intersection function $\mathbb{I}$ (please refer to its illustration provided in Appendix.~\ref{appendix:Model Architecture}) is built based on the encoder architecture of vision transformer~\cite{dosovitskiy2020image} (ViT)
, in which its input is the embeddings of the seen attributes, i.e. the \rebuttal{model} takes $m^s_k$ and $m^s_l$ from $M^s$ as input when performing $\mathbb{I}(a_k, a_l)$, where $a_k, a_l \in \mathbf{A}^s$. 
\walon{To be detailed, there are several modifications in our \rebuttal{model} for intersection $\mathbb{I}$ with respect to the original ViT: (1) We remove the position embedding in order to fulfil the commutative property of intersection, i.e. $\mathbb{I}(a_k, a_l) = \mathbb{I}(a_l, a_k)$; (2) We attach a learnable token named ``intersection head'' to the input sequence of transformer, which is similar to the extra class embedding in ViT. The corresponding output of this intersection head after going through the transformer encoder represents the embedding of the resultant base attribute, where we apply the element-wise absolute-value operation on it to make it a non-negative vector (being analogous to what we did for the seen attributes)\rebuttal{; (3) There is only a single self-attention block in the transformer}. Please note that, the embedding of a base attribute is also a $C$-dimensional vector. Regarding our union function $\mathbb{U}$, we simply adopt the average operation for its implementation, that is: Given two base attributes $b^c \in \mathbf{B}^c$ and $b^p \in \mathbf{B}^p$, we obtain the embedding $\tilde{m}$ of the synthesized attribute $\tilde{a} = \mathbb{U}(b^c, b^p)$ by averaging the embeddings of $b^c$ and $b^p$. Specifically, such $C$-dimensional embedding $\tilde{m}$ is also defined upon the image feature and acts as the detector for the synthesized attribute $\tilde{a}$. }

\walon{The training of our decompose-and-reassemble procedure for synthesizing novel attributes is simply based on the reconstruction loss of the seen attributes $\mathcal{L}_{rec}$. Given a synthesized attribute $\tilde{a}$, if there exists a seen attribute $a_k \in \mathbf{A}^s $ with having $\beta^c(a_k) = \beta^c(\tilde{a})$ and $\beta^p(a_k) = \beta^p(\tilde{a})$, the embedding $\tilde{m}$ of $\tilde{a}$ and the embedding $m^s_k$ of $a_k$ are expected to be identical to each other, $\mathcal{L}_{rec}$ is thus defined as:
\begin{equation}
    \mathcal{L}_{rec} = \left \| m^s_k - \tilde{m} \right\|
    \label{eq:rec}
\end{equation}
Note that, as our union function $\mathbb{U}$ has no trainable parameters (since it is simply an average operation), the gradient of $\mathcal{L}_{rec}$ is propagated to focus on learning the parameters of our transformer for the intersection function $\mathbb{I}$. In other words, we expect that the transformer is so powerful to be capable of extracting the base attributes where their averages are informative enough to act as the detectors for the synthesized attributes. Note that in order to fully leverage the seen attributes for training our decompose-and-reassemble procedure, we have a particular training scheme where its algorithm and the implementation details are shown in the Appendix.~\ref{appendix:algorithm}.}

\tocless\section{Experimental Results}
\label{sec:exp_result}

\noindent\textbf{Dataset.}
Our experiments are mainly conducted on the Caltech-UCSD Birds-200-2011 dataset~\cite{CUB} (usually abbreviated as CUB) for zero-shot classification. CUB dataset collects 11,788 images of 200 bird categories, where each image is annotated with 312 attributes. We select 32 attributes as our seen attributes $\mathbf{A}^s$, which can be decomposed into 15 base attributes of adjective $\mathbf{B}^c$ and 16 base attributes of object part $\mathbf{B}^p$, and we use these base attributes to synthesize 207 novel attributes $\mathbf{A}^u$.
We follow the setting proposed by~\cite{ZSLGBU} for the task of generalized zero-shot learning (GZSL) to split the CUB dataset, where such training and testing sets are used to train and evaluate our proposed scenario of ZSL on attributes, respectively.
\modify{
Furthermore, we base on~\cite{clevr} to create a synthetic dataset, named $\alpha$-CLEVR, for providing extensive analysis on the robustness of our proposed ZSLA against the noisy seen attributes (which are related to an issue hidden behind CUB dataset).}
\modify{$\alpha$-CLEVR contains 24 attributes (i.e. types of toy bricks) which are the combinations of 8 colors (i.e., base attributes of adjective $\mathbf{B}^c$) and 3 shapes (i.e., base attributes of object part $\mathbf{B}^p$). Among them, 16 attributes, which can be decomposed into the 11 base attributes, are selected as seen attributes $\mathbf{A}^s$ for training our proposed ZSLA, where we are then able to synthesize novel attribute detectors to annotate the images of $\alpha$-CLEVR dataset. Noting that, classes in our $\alpha$-CLEVR dataset are defined by the specific combinations of toy bricks (where toy bricks with different color-shape combinations are treated as different attributes)
For performing the GZSL task on $\alpha$-CLEVR and evaluate the annotation quality of our synthesized attribute detectors, we create 160 classes in $\alpha$-CLEVR, where each class has 30 images; 80 classes are set as seen data, and the other 80 are set as unseen data. 
For GZSL inference, testing images from both seen and unseen classes are used.} 


\begin{table*}[t]
\centering
\caption{Evaluation of synthesized novel/unseen attributes on attribute classification (mAUROC), retrieval (mAP@50), and localization (mLA). $N^s$ is the number of seen attributes.}
\vspace{.5em}
\resizebox{1.0\textwidth}{!}{
\begin{tabular}{c|ccc|ccc|ccc}
                  & \multicolumn{3}{c|}{\textbf{mAUROC}}                                                                                  & \multicolumn{3}{c|}{\textbf{mAP@50}}                                                                                     & \multicolumn{3}{c}{\textbf{mLA}}                                                                                  \\ \hline
\textbf{$N^s$}       & {\color[HTML]{3531FF} 32}             & {\color[HTML]{009901} 64}             & {\color[HTML]{CE6301} 96}             & {\color[HTML]{3531FF} 32}             & {\color[HTML]{009901} 64}             & {\color[HTML]{CE6301} 96}             & {\color[HTML]{3531FF} 32}             & {\color[HTML]{009901} 64}             & {\color[HTML]{CE6301} 96}             \\ \hline
\textbf{A-LAGO}   & {\color[HTML]{3531FF} 0.600}          & {\color[HTML]{009901} 0.612}          & {\color[HTML]{CE6301} 0.627}          & {\color[HTML]{3531FF} 0.173}          & {\color[HTML]{009901} 0.180}          & {\color[HTML]{CE6301} 0.222}          & {\color[HTML]{3531FF} 0.782}          & {\color[HTML]{009901} 0.787}          & {\color[HTML]{CE6301} 0.795}          \\
\textbf{A-ESZSL}  & {\color[HTML]{3531FF} 0.626}          & {\color[HTML]{009901} 0.614}          & {\color[HTML]{CE6301} 0.632}          & {\color[HTML]{3531FF} 0.223}          & {\color[HTML]{009901} 0.200}          & {\color[HTML]{CE6301} 0.234}          & {\color[HTML]{3531FF} 0.756}          & {\color[HTML]{009901} 0.769}          & {\color[HTML]{CE6301} 0.756}          \\
\textbf{Our ZSLA} & {\color[HTML]{3531FF} \textbf{0.689}} & {\color[HTML]{009901} \textbf{0.704}} & {\color[HTML]{CE6301} \textbf{0.717}} & {\color[HTML]{3531FF} \textbf{0.320}} & {\color[HTML]{009901} \textbf{0.327}} & {\color[HTML]{CE6301} \textbf{0.329}} & {\color[HTML]{3531FF} \textbf{0.846}} & {\color[HTML]{009901} \textbf{0.860}} & {\color[HTML]{CE6301} \textbf{0.867}}
\end{tabular}
}
\label{tab:table1}
\end{table*}

\noindent\textbf{Baselines.}
As our task of ZSL on attributes for dataset annotation is novel, there is no prior work that we can directly make a comparison with. \modify{However, as the task of ZSL on attribute has a hierarchy between attributes and base attributes,} we adapt two representative methods of zero-shot classification (which explicitly have the class--attribute hierarchy behind their formulation) to be our baselines, by using the analogy between two hierarchies (i.e. our attribute--base attribute versus their class--attribute). These two baselines are ESZSL~\cite{ESZSL} and LAGO singleton~\cite{LAGO} (note that both of them realize classification with the help of attribute prediction), in which we particularly rename their adaptions to our scenario of ZSL on attributes as $\mathbf{A}$-\textbf{ESZSL} and $\mathbf{A}$-\textbf{LAGO} respectively for avoiding confusion. \modify{ Details of implementing baselines are provided in \CR{Appendix \ref{appendix:baseline}.}}

Note that, in the following experiments, both the baselines and ZSLA use the additional ground-truth of attribute locations (i.e. knowing where an attribute appears on the image) provided by CUB to train the seen attribute detectors, unless stated otherwise.

\nips{\textbf{Limitation and discussion.} Our proposed method (ZSLA) tackles zero-shot learning for unseen attributes and enables the efficient attribute annotations when constructing the new datasets (later shown in Sec~\ref{sec:reannotate}) as long as the attribute format can be re-factored (e.g. blue wing) for assembling novel attributes via the decompose-and-reassemble approach. Although such requirement on the attribute format restricts the direct application of our ZSLA on some ZSL datasets (e.g. AWA2 and SUN datasets), the particular attribute format adopted in our ZSLA actually better fits the human's intuition for describing the discriminative attributes for the ``fine-grained'' classes. Hence, we would not see this format requirement on attributes as the limitation of our proposed method but the restriction of the mainstream ZSL datasets for not having the fine-grained attributes.}

\subsection{Evaluation of Unseen Attributes}
\label{sec:experiment1}

We design three schemes to evaluate the quality of the synthesized novel attribute detectors learnt by ZSLA: (1) \textbf{Attribute Classification.} Based on ground-truth attribute annotation of the test images (note that each image typically has multiple attributes), we measure the performance of our synthesized attribute detectors on recognizing their corresponding attributes in the test images. We adopt the area under receiver operating characteristic (AUROC) as our metric for the classification accuracy of each attribute, and we report the average over AUROCs (denoted as mAUROC) of all synthesized attribute detectors; (2) \textbf{Attribute Retrieval.} We rank the test images according to their image-wise responses as to a given attribute detector, to simulate the application scenario of retrieving the images which are most likely to own the target attribute from an image set. Note that the image-wise response is computed by max-pooling over the responses of patch-wise image features with respect to the attribute detector. For each attribute detector we compute the average precision (AP) of its top 50 retrieved images, and report the average AP (denoted as mAP@50) of all detectors as the metric; (3) \textbf{Attribute Localization.} As in CUB the ground-truth locations that an attribute appears on the test images are available, we introduce the localization accuracy (LA) to measure how well the location having the highest response to an attribute detector matches with the ground-truth ones (counted as correct if they are located on the same or neighboring patches). We average over the LA of each attribute as the metric (denoted as mLA).
\begin{figure*}[t] 
\centering 
\includegraphics[width=1\textwidth]{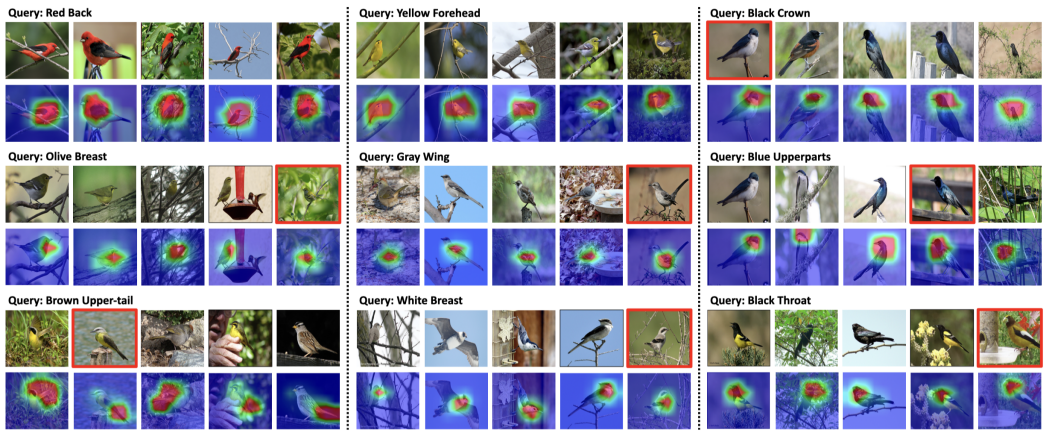} 
\vspace{-1.5em}
\caption{Examples of attribute retrieval and localization. Each set shows the top-5 retrieved images and their response maps for a synthesized novel attribute, where the images marked with red borders are the false positives according to CUB ground-truth.
}
\label{retrieval} 
\end{figure*}
\begin{figure*}[t]
    \centering
    \includegraphics[width=1\textwidth]{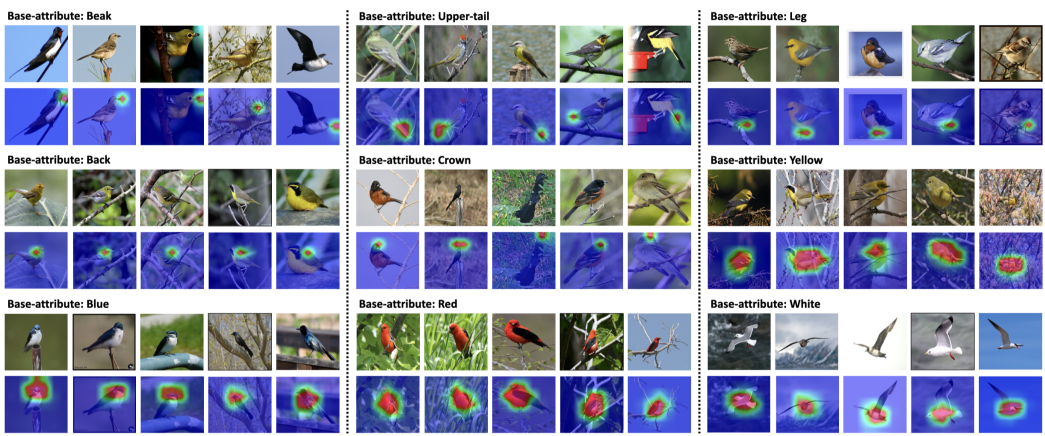}
    \vspace{-1.5em}
    \caption{Examples of showing the retrieval and localization ability of base attributes. Each set shows the top-5 retrieved images and their corresponding response map for a base attribute representation (extracted by applying our intersection operation on seen attributes detectors).}
    \label{Fig.base_retrieval}
\end{figure*}

Table~\ref{tab:table1} summarizes the performance in terms of mAUROC, mAP@50, and mLA obtained by baselines and ZSLA, with the number $N^s$ of seen attributes $\mathbf{A}^s$ set as $\{32, 64, 96\}$. It is clear to see that ZSLA provides superior performance in comparison to the baselines on all the settings of $N^s$ and evaluation schemes, particularly the localization accuracy. Moreover, by using merely 32 seen attributes to perform the synthesis of novel attribute detectors, ZSLA can achieve comparable results with the baselines of using 64 or 96 seen attributes. Qualitative examples for showing the results of attribute retrieval and attribute localization for the novel attributes synthesized by ZSLA are provided in Figure~\ref{retrieval}.
Besides these quantitative and qualitative results demonstrating the efficacy of ZSLA on novel attributes, we also provide some qualitative examples in Figure~\ref{Fig.base_retrieval} to showcase the localization and retrieval ability of our base attribute representations extracted from the seen attribute detectors.

\subsection{Automatic Annotations for Learning Generalized Zero-Shot Image Classification}
\label{sec:reannotate}
To further access the quality of our synthesized attribute detectors, we adopt the 32 seen attribute detectors and the 207 novel attribute detectors (i.e. $N^s$=32, $N^u$=207) learned by ZSLA to \textit{re-annotate the attribute labels for the whole CUB dataset} to simulate the labeling process during constructing a new dataset, and name the resultant new dataset ``$\delta$-CUB''. Then we adopt $\delta$-CUB to train and evaluate four representative 
GZSL algorithms, i.e. ALE~\cite{ALE}, ESZSL~\cite{ESZSL}, CADAVAE~\cite{CADAVAE}, and TFVAEGAN~\cite{tfVAEGAN} using the settings proposed by~\cite{ZSLGBU} (i.e. for $\delta$-CUB and CUB, training with samples from the 150 seen classes, then evaluating the performance on all 200 classes including the 50 unseen ones). Note that, the class-attribute matrix, which shows the composition of attributes for each class and is needed for GZSL (i.e. the semantic information of classes), is computed by the statistics in $\delta$-CUB. Similarly, we also use $\mathbf{A}$-\textbf{ESZSL} and $\mathbf{A}$-\textbf{LAGO} baselines to re-annotate CUB dataset and perform GZSL under the same aforementioned setting. The results related to ZSLA and baselines are summarized in the row shaded by the orange color of Table~\ref{tab:deltaCUB_GZSL}. 
Moreover, we additionally experiment on training the four GZSL algorithms by using only 32 attributes or using all 312 attributes obtained from the original CUB dataset as the semantic information, where their results are summarized in the rows shaded by the blue and green color of Table~\ref{tab:deltaCUB_GZSL}, respectively.

\new{
From the results, we observe that using $\delta$-CUB for training, where our ZSLA automatically annotates all the attribute labels, can largely benefit the performance of GZSL algorithms. By treating the harmonic mean over the accuracy numbers on both seen and unseen categories as the metric for GZSL, $\delta$-CUB is superior to those datasets annotated by baselines or even the one using manual annotations.
Specifically, the gain obtained by using our $\delta$-CUB with respect to the setting of using 32 manually-labeled attributes (i.e., the blue-shaded row of Table~\ref{tab:deltaCUB_GZSL}) demonstrates the practical value of our proposed problem scenario of ZSL on attributes: Without additional cost for collecting annotation, we provide more attribute labels via synthesizing novel attribute detectors from the seen ones, and thus different categories can be better distinguished by more fine-grained/detailed attribute-based representations. 
Moreover, regarding the results that our automatic re-annotation leads to better performance than the manual one (i.e., the green-shaded row of Table~\ref{tab:deltaCUB_GZSL}), we believe that this is mainly due to the biased semantic information caused by noisy labels stemming from the inconsistency between different human annotators when building CUB dataset. In comparison, our attribute detectors can produce consistent attribute annotations as we use the same set of attribute detectors for labeling all images; it eventually contributes to a more suitable semantic for learning zero-shot classification. We provide more discussions on such issues in \CR{Appendix~\ref{appendix:label consistency problem}}}

\begin{table*}[t!]
\caption{
Experiments results of training and evaluating four representative GZSL methods (i.e. CADAVAE, TFVAEGAN, ALE, ESZSL) on the datasets built upon different sources of attribute annotation (e.g. manual annotation given by original CUB dataset, and re-annotation provided by ZSLA or baselines).
As for the columns, $\textbf{S}$ and $\textbf{U}$ represent the accuracy on seen and unseen classes respectively, while $\textbf{H}$ represents the harmonic mean of $\textbf{S}$ and $\textbf{U}$. The highest scores are marked in bold \textcolor{red}{red}, while the second-highest ones are marked in bold \textcolor{blue}{blue}. 
Particularly, we encourage the readers to observe the relative improvement/gain (in terms of harmonic mean) produced by various approaches/settings with respect to the results obtained by using 32 manually-labelled seen attributes for GZSL (i.e. the results on the blue-shaded row for CUB dataset), where the superior gain contributed by our propose ZSLA well demonstrates its practical value of automatically producing high-quality annotations on the novel attributes.  
}
\vspace{0.5em}
\resizebox{\textwidth}{!}{
\begin{tabular}{c|ccc|ccc|ccc|ccc|}
\cline{2-13}
                                                                                     & \multicolumn{3}{c|}{CADAVAE~\cite{CADAVAE}}                                                                                                 & \multicolumn{3}{c|}{TFVAEGAN~\cite{tfVAEGAN}}                                                                                               & \multicolumn{3}{c|}{ALE~\cite{ALE}}                                                                                                         & \multicolumn{3}{c|}{ESZSL~\cite{ESZSL}}                                                                                                     \\ \cline{2-13} 
                                                                                     & S                                            & U                                            & H                                             & S                                            & U                                            & H                                             & S                                            & U                                            & H                                             & S                                            & U                                            & H                                             \\ \hline
\rowcolor[HTML]{C9DAF8} 
\multicolumn{1}{|c|}{\cellcolor[HTML]{C9DAF8}Manual}                                 & 42.9                                         & 27.3                                         & 33.4                                          & 45.5                                         & 31.2                                         & 37.1                                          & 26.4                                         & 9.2                                          & 13.7                                          & 29.8                                         & 10.8                                         & 15.9                                          \\
\rowcolor[HTML]{C9DAF8} 
\multicolumn{1}{|c|}{\cellcolor[HTML]{C9DAF8}($N^s$=32 for CUB)}                     & \multicolumn{1}{l}{\cellcolor[HTML]{C9DAF8}} & \multicolumn{1}{l}{\cellcolor[HTML]{C9DAF8}} & \multicolumn{1}{l|}{\cellcolor[HTML]{C9DAF8}} & \multicolumn{1}{l}{\cellcolor[HTML]{C9DAF8}} & \multicolumn{1}{l}{\cellcolor[HTML]{C9DAF8}} & \multicolumn{1}{l|}{\cellcolor[HTML]{C9DAF8}} & \multicolumn{1}{l}{\cellcolor[HTML]{C9DAF8}} & \multicolumn{1}{l}{\cellcolor[HTML]{C9DAF8}} & \multicolumn{1}{l|}{\cellcolor[HTML]{C9DAF8}} & \multicolumn{1}{l}{\cellcolor[HTML]{C9DAF8}} & \multicolumn{1}{l}{\cellcolor[HTML]{C9DAF8}} & \multicolumn{1}{l|}{\cellcolor[HTML]{C9DAF8}} \\
\rowcolor[HTML]{DFF8DE} 
\multicolumn{1}{|c|}{\cellcolor[HTML]{DFF8DE}Manual}                                 & {\color[HTML]{FE0000} \textbf{53.5}}         & 51.6                                         & 52.4                                          & {\color[HTML]{FE0000} \textbf{64.7}}         & 52.8                                         & {\color[HTML]{FE0000} \textbf{58.1}}          & {\color[HTML]{FE0000} \textbf{62.8}}         & 23.7                                         & 34.4                                          & {\color[HTML]{3531FF} \textbf{63.8}}         & 12.6                                         & 21.0                                          \\
\rowcolor[HTML]{DFF8DE} 
\multicolumn{1}{|c|}{\cellcolor[HTML]{DFF8DE}($N^s$=312 for CUB)}                    & \multicolumn{1}{l}{\cellcolor[HTML]{DFF8DE}} & \multicolumn{1}{l}{\cellcolor[HTML]{DFF8DE}} & \multicolumn{1}{l|}{\cellcolor[HTML]{DFF8DE}} & \multicolumn{1}{l}{\cellcolor[HTML]{DFF8DE}} & \multicolumn{1}{l}{\cellcolor[HTML]{DFF8DE}} & \multicolumn{1}{l|}{\cellcolor[HTML]{DFF8DE}} & \multicolumn{1}{l}{\cellcolor[HTML]{DFF8DE}} & \multicolumn{1}{l}{\cellcolor[HTML]{DFF8DE}} & \multicolumn{1}{l|}{\cellcolor[HTML]{DFF8DE}} & \multicolumn{1}{l}{\cellcolor[HTML]{DFF8DE}} & \multicolumn{1}{l}{\cellcolor[HTML]{DFF8DE}} & \multicolumn{1}{l|}{\cellcolor[HTML]{DFF8DE}} \\
\rowcolor[HTML]{F8E8D5} 
\multicolumn{1}{|c|}{\cellcolor[HTML]{F8E8D5}A-LAGO}                                 & 45.4                                         & {\color[HTML]{3531FF} \textbf{55.4}}         & 49.9                                          & 57.4                                         & {\color[HTML]{3531FF} \textbf{53.0}}         & 55.1                                          & 51.8                                         & {\color[HTML]{3531FF} \textbf{27.2}}         & {\color[HTML]{3531FF} \textbf{35.6}}          & 49.7                                         & {\color[HTML]{FE0000} \textbf{17.1}}         & {\color[HTML]{3531FF} \textbf{25.4}}          \\
\rowcolor[HTML]{F8E8D5} 
\multicolumn{1}{|c|}{\cellcolor[HTML]{F8E8D5}A-ESZSL}                                & 41.5                                         & 48.7                                         & 44.8                                          & 56.0                                         & 48.5                                         & 52.0                                          & 49.7                                         & 17.1                                         & 25.4                                          & 61.3                                         & 9.2                                          & 16.0                                          \\
\rowcolor[HTML]{F8E8D5} 
\multicolumn{1}{|c|}{\cellcolor[HTML]{F8E8D5}Our ZSLA}                               & {\color[HTML]{3531FF} \textbf{50.3}}         & {\color[HTML]{FE0000} \textbf{56.4}}         & {\color[HTML]{FE0000} \textbf{53.2}}          & {\color[HTML]{3531FF} \textbf{59.0}}         & {\color[HTML]{FE0000} \textbf{55.9}}         & {\color[HTML]{3531FF} \textbf{57.4}}          & {\color[HTML]{3531FF} \textbf{52.4}}         & {\color[HTML]{FE0000} \textbf{27.5}}         & {\color[HTML]{FE0000} \textbf{36.1}}          & {\color[HTML]{FE0000} \textbf{65.1}}         & {\color[HTML]{3531FF} \textbf{16.4}}         & {\color[HTML]{FE0000} \textbf{26.2}}          \\
\rowcolor[HTML]{F8E8D5} 
\multicolumn{1}{|c|}{\cellcolor[HTML]{F8E8D5}($N^s$=32, $N^u$=207 for $\delta$-CUB)} & \multicolumn{1}{l}{\cellcolor[HTML]{F8E8D5}} & \multicolumn{1}{l}{\cellcolor[HTML]{F8E8D5}} & \multicolumn{1}{l|}{\cellcolor[HTML]{F8E8D5}} & \multicolumn{1}{l}{\cellcolor[HTML]{F8E8D5}} & \multicolumn{1}{l}{\cellcolor[HTML]{F8E8D5}} & \multicolumn{1}{l|}{\cellcolor[HTML]{F8E8D5}} & \multicolumn{1}{l}{\cellcolor[HTML]{F8E8D5}} & \multicolumn{1}{l}{\cellcolor[HTML]{F8E8D5}} & \multicolumn{1}{l|}{\cellcolor[HTML]{F8E8D5}} & \multicolumn{1}{l}{\cellcolor[HTML]{F8E8D5}} & \multicolumn{1}{l}{\cellcolor[HTML]{F8E8D5}} & \multicolumn{1}{l|}{\cellcolor[HTML]{F8E8D5}} \\ \hline
\end{tabular}
}
\label{tab:deltaCUB_GZSL}
\end{table*}

\subsection{Robustness Against Noisy Attribute Labels}\label{sec:robustness}

\new{
Due to the preference bias among different annotators mentioned in Section 4.2, it is hard to obtain perfect seen attribute labels for training. Thus, it is interesting to discuss the effect of the noisy level of seen attribute labels (used for training) on the final annotation quality produced by different auto-annotation methods.}
\modify{ Thus, we conduct the controlled experiments using $\alpha$-CLEVR to understand the effect of the noisy labels.}
\modify{In detail, we adjust the amount of noisy labels for the analysis.} \new{To measure the performance drop caused by noisy seen attribute labels, we define the \textbf{wrong attribute label rate} (abbreviated as \textbf{\textbf{\textbf{\textbf{WALR}}}}) to represent the noisy level of attribute labels. 
For example, when WALR is set to 0.3, any toy brick in the training images has a 30\% chance of inaccurately annotating (e.g., a blue cube is annotated as a red sphere). 
}

\begin{figure}[t!]
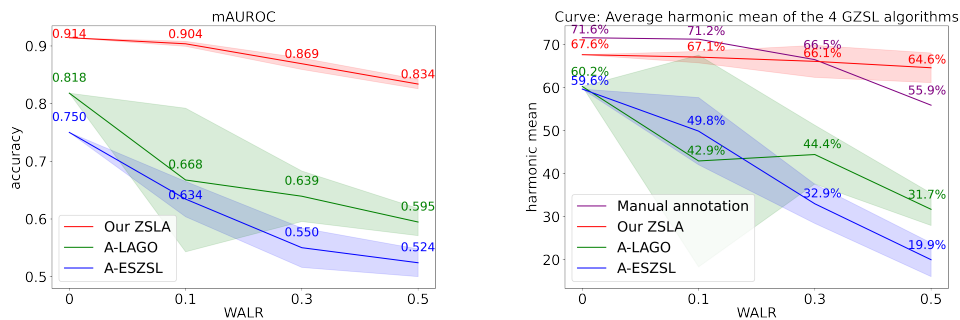

\centering
\subfigure{
\label{Fig.noisy_label_95_CI}
\includegraphics[width=0.47\textwidth]{CI_mAUROC.pdf}
}
\subfigure{
\label{Fig.noisy_label_GZSLAH}
\includegraphics[width=0.47\textwidth]{ClevrGZSLAvgH.pdf}
}
\caption{\nips{(a) Left: Evaluation (in terms of attribute classification, with mAUROC as the metric) on the robustness against noisy attribute labels for various methods which learn to synthesize the novel attributes. (b) Right: Evaluation on the quality of automatic re-annotation produced by different methods, where the performance is based on the average harmonic mean of four GZSL algorithms using the re-annotated attributes (cf. the last paragraph of Sec.~\ref{sec:robustness} for details). Noting that each curve reports the average\ performance over 5 runs of different noisy label sets (as the label noise is randomly injected), while the shaded bands around each curve represent the 95\% confidence interval.
}}
\label{fig:figures}
\end{figure}

\new{ 
As the mAUROC curves shown in Figure~\ref{Fig.noisy_label_95_CI}, we can observe that: (1) ZSLA outperforms the baselines in attribution classification, no matter how noisy the training data is; (2) the performance drop of ZSLA with respect to WALR is much smaller than that of the baselines; (3) baselines have a larger variance than ours, i.e., they are more sensitive to different combinations of the noisy labels even the noise level is the same. The observations prove the robustness of ZSLA against the noisy labels of training attributes. We also show in \CR{Figure \ref{Fig.CI_mAP_at_50_mLA} in Appendix} that mAP@50 and mLA (for attribute retrieval and location) have a similar trend as mAUROC.}

\walon{Moreover, similar to the experimental setting as Section~\ref{sec:reannotate}, we use the novel attribute detectors, 
synthesized by different methods under various WALR settings, to automatically re-annotate the dataset. The resultant dataset is used for learning four GZSL algorithms (i.e., CADAVAE, TFVAEGAN, ALE, and ESZSL). The average of their harmonic means is reported in Figure~\ref{Fig.noisy_label_GZSLAH}. We can observe the superior quality in terms of automatic re-annotation produced by our ZSLA (i.e., the red curve) compared to the other baselines (i.e., the blue and green curves for A-ESZSL A-LAGO, respectively) under all WALR settings. Specifically, we also simulate the situation where humans annotate all attribute labels for the dataset while maintaining the corresponding WALRs (i.e., the purple curve).  It leads to a similar observation as we find in the CUB dataset. Once WALR is high (i.e., quite noisy labeling), the performance of GZSL algorithms trained with the semantic information provided by our ZSLA (i.e., the red curve) becomes superior to the one trained with the noisy manual labels.
}


\tocless
\section{Conclusion}
 \new{This paper proposes a new method of developing zero-shot learning on novel attributes to reduce the attribute annotation cost for constructing a zero-shot classification dataset. By leveraging the trained detectors of seen attributes, our model learns to decompose them into base attributes to further synthesize novel unseen attributes by reassembling pairs of base attributes.
 Experimental results show that our method is able to exploit the information embedded in the seen attributes to generate high-quality unseen attributes, validated by various evaluation schemes for attribute classification, retrieval, and localization. We also demonstrate that the semantic information based on our automatic re-annotation is beneficial for the GZSL task.\\}

\textbf{Acknowledgement.} This project is supported by NSTC (National Science and Technology Council, Taiwan) 111-2636-E-A49-003, 111-2628-EA49-018-MY4, 110-2221-E-A49-066-MY3 and 109-2221-E-009-112-MY3. We are grateful to the National Center for High-performance Computing for computer time and facilities.

\bibliographystyle{plainnat}
\bibliography{6_reference}
\newpage
\tocless
\section*{Checklist}

The checklist follows the references.  Please
read the checklist guidelines carefully for information on how to answer these
questions.  For each question, change the default \answerTODO{} to \answerYes{},
\answerNo{}, or \answerNA{}.  You are strongly encouraged to include a {\bf
justification to your answer}, either by referencing the appropriate section of
your paper or providing a brief inline description.  For example:
\begin{itemize}
  \item Did you include the license to the code and datasets? \answerYes{} 
  \item Did you include the license to the code and datasets? \answerNo{The code and the data are proprietary.}
  \item Did you include the license to the code and datasets? \answerNA{}
\end{itemize}
Please do not modify the questions and only use the provided macros for your
answers.  Note that the Checklist section does not count towards the page
limit.  In your paper, please delete this instructions block and only keep the
Checklist section heading above along with the questions/answers below.

\begin{enumerate}

\item For all authors...
\begin{enumerate}
  \item Do the main claims made in the abstract and introduction accurately reflect the paper's contributions and scope?
    \answerYes{}
  \item Did you describe the limitations of your work?
    \answerYes{See the ``\textbf{Limitation and discussion}" part in Section~\ref{sec:exp_result}.}
  \item Did you discuss any potential negative societal impacts of your work?
    \answerNo{To the best of our understanding, there are no potential negative societal impacts in our work.}
  \item Have you read the ethics review guidelines and ensured that your paper conforms to them?
    \answerYes{}
\end{enumerate}

\item If you are including theoretical results...
\begin{enumerate}
  \item Did you state the full set of assumptions of all theoretical results?
    \answerNA{}
        \item Did you include complete proofs of all theoretical results?
    \answerNA{}
\end{enumerate}

\item If you ran experiments...
\begin{enumerate}
  \item Did you include the code, data, and instructions needed to reproduce the main experimental results (either in the supplemental material or as a URL)?
    \answerYes{Please refer to our supplementary materials.}
  \item Did you specify all the training details (e.g., data splits, hyperparameters, how they were chosen)?
    \answerYes{Please refer to our code and our supplementary materials.}
        \item Did you report error bars (e.g., with respect to the random seed after running experiments multiple times)?
    \answerYes{For the experiment result that strongly impacted by the random sampling process, we repeat the experiment for several rounds and report the mean performance as well as the 95\% confident region. See Section~\ref{sec:robustness}.}
        \item Did you include the total amount of compute and the type of resources used (e.g., type of GPUs, internal cluster, or cloud provider)?
    \answerNo{}
\end{enumerate}

\item If you are using existing assets (e.g., code, data, models) or curating/releasing new assets...
\begin{enumerate}
  \item If your work uses existing assets, did you cite the creators?
    \answerYes{See the ``\textbf{Dataset}" part in Section~\ref{sec:exp_result}. }
  \item Did you mention the license of the assets?
    \answerNo{}
  \item Did you include any new assets either in the supplemental material or as a URL?
    \answerNo{}
  \item Did you discuss whether and how consent was obtained from people whose data you're using/curating?
    \answerNo{We use the public dataset with referencing to the original paper/resource.}
  \item Did you discuss whether the data you are using/curating contains personally identifiable information or offensive content?
    \answerNo{To the best of our understanding, there are no related issue in the applied dataset.}
\end{enumerate}

\item If you used crowdsourcing or conducted research with human subjects...
\begin{enumerate}
  \item Did you include the full text of instructions given to participants and screenshots, if applicable?
    \answerNA{}
  \item Did you describe any potential participant risks, with links to Institutional Review Board (IRB) approvals, if applicable?
    \answerNA{}
  \item Did you include the estimated hourly wage paid to participants and the total amount spent on participant compensation?
    \answerNA{}
\end{enumerate}

\end{enumerate}
\newpage
\section*{Appendix}
\tableofcontents
\newpage
\setcounter{subsection}{0}
\renewcommand{\thesubsection}{\Alph{subsection}}
\subsection{Supplementary for Proposed Method}
\subsubsection{Training Scheme of Decompose-and-Reassemble Algorithm}
\label{appendix:algorithm}
\appendixNew{As discussed in Section~3.2 
of the main manuscript, we propose a training scheme to train our intersection module $\mathbb{I}$ such that it can well decompose the common concepts between (seen) attributes for obtaining the base attributes. The workflow of our proposed training scheme is as follows:}

\begin{algorithm}[!htp]
\SetAlgoLined
\SetKwInput{KwData}{Given}
\KwData{trained detectors $M^s$ of seen attributes $\mathbf{A}^s$}
\KwResult{parameters $\theta$ of the transformer for $\mathbb{I}$}
 \For{\text{every attribute } $a \in \mathbf{A}^s$}{
  randomly sample attributes $a_k$, $a_l$ from $\mathbf{A}^s$ with\\ $\beta^c(a)=\beta^c(a_k)=\beta^c(a_l)$,  $\beta^p(a_k)\neq\beta^p(a_l)$;\\
  obtain the embedding $m^c$ of base attribute $\beta^c(a)$ via intersection $\mathbb{I}(a_k, a_l)$;\\
  randomly sample attributes $a_{{k}'}$, $a_{{l}'}$ from $\mathbf{A}^s$ with\\ $\beta^p(a)=\beta^p(a_{{k}'})=\beta^p(a_{{l}'})$, $\beta^c(a_{{k}'})\neq\beta^c(a_{{l}'})$;\\
  obtain the embedding $m^p$ of base attribute $\beta^p(a)$ via intersection $\mathbb{I}(a_{{k}'}, a_{{l}'})$;\\
  synthesize attribute $\tilde{a}$ via union $\mathbb{U}(\beta^c(a), \beta^p(a))$ with its embedding $\tilde{m} = (1/2) \cdot (m^c+m^p)$;\\
  $\theta \leftarrow \arg\min\limits_{\theta} \mathcal{L}_{rec}(m^s_k, \tilde{m})$;
 }
\caption{\appendixNew{Decompose-and-Reassemble}}
\end{algorithm}

\par\appendixNew{
Recap that: (1) $\mathbf{A}^s$ is the seen attribute set; (2) $\mathbf{B}^c$ and $\mathbf{B}^p$ are the base attribute sets of the adjectives and object parts respectively; (3) $\beta^c(a)$ and $\beta^p(a)$ are the corresponding base attributes respectively of the adjective part and the object part for a given attribute $a$~(i.e. $\beta^c(a) \in \mathbf{B}^c$ and $\beta^p(a) \in \mathbf{B}^p$); (4) we denote the intersection operation as $\mathbb{I}$ and the union operation as $\mathbb{U}$, where $\mathbb{I}$ is built as a neural network (specifically based on the encoder architecture of vision transformer~\cite{dosovitskiy2020image}) with parameters $\theta$, and $\mathbb{U}$ adopts a simple average function; (5) $m^s_k$ denotes the embedding of the seen attribute $a_k$. 
}

\subsubsection{Model Architecture}
\label{appendix:Model Architecture}
\begin{figure}[h] 
\centering 
\includegraphics[width=0.85\columnwidth]{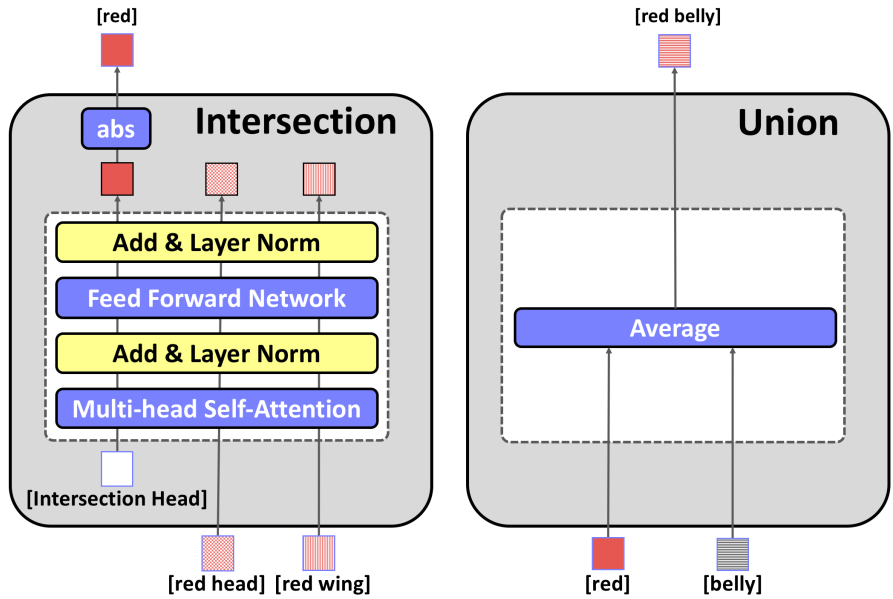} 
\vspace{-0.5em}
\caption{\walon{The implementation of our intersection $\mathbb{I}$ and union $\mathbb{U}$ operations to realize the decompose-and-reassemble procedure, where $\mathbb{I}$ adopts the architecture extended from the vision transformer~\cite{dosovitskiy2020image} while $\mathbb{U}$ simply adopts the average operation. }}
\label{Fig.architecture} 
\end{figure}
\subsubsection{Implementation Details} 
\label{appendix:implementation_details}
\modify{
To train the seen attribute detectors $M^s$, we use the Adam optimizer with a learning rate of $10^{-3}$, weight decay of $10^{-4}$, and beta values of $(0.5, 0.9)$. We adopt the ImageNet-pretrained ResNet101 network as our feature extractor $f$, in which the feature map extracted by $f$ has $C=2048$ channels. Although the feature extractor $f$ can be jointly trained with the detectors $M^s$ in our proposed framework, we choose to keep it fixed to follow the common setting in~\cite{ZSLGBU}. For our intersection operation $\mathbb{I}$, it has one transformer block and 16 heads in its multi-head attention layer; the dimension of each head is 64. To train the transformer, we use Adam optimizer with a learning rate of $10^{-4}$, weight decay of $10^{-4}$, beta values of $(0.5, 0.9)$, and a dropout rate of $10^{-1}$. 
Moreover, as cosine similarity is used to compute attribute response map, attribute detectors $M^s$, $M^n$ and their base attributes are L2-normalized during training; thus, the model does not need to care about their scale of vectors.}

\subsubsection{The Assumption of Uni-modal Constraint}
\label{appendix:UMC}
\begin{figure}[h!]
    \centering
    \includegraphics[width=0.40\textwidth]{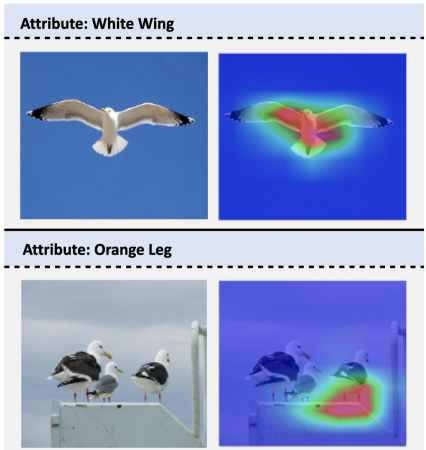}
    \vspace{-1em}
    \caption{
    \modify{Examples and their response maps for the attribute that appear at multiple locations: a California gull with spread wings~(top image) and a family of slaty backed gulls~(bottom image).}}
    \label{fig:umc_sample}
    \vspace{-1em}
\end{figure}
\modify{
During the training of seen attribute detectors, we utilize uni-modal constraint to encourage the single peak of attribute response map assuming that an attribute only appears at only one location or a small region on the image. Such an assumption might not be true for all cases. For instance, as shown in Figure~\ref{fig:umc_sample}, a flying bird has spread wings (i.e. left-wing and right-wing) located far from each other; or birds of one species in a single image have duplicated attributes occurring at different places. Although the assumption behind the uni-modal constraint is sometimes violated, an attribute detector is able to work well even if it is forced to focus on only one location. That is, ``white wing'' detector can learn what is ``white wing'' no matter by the left-wing or the right-wing~(top image in Figure~\ref{fig:umc_sample}); similarly, the detector can get the concept of ``orange leg'' by seeing any bird with that attribute~(bottom image in Figure~\ref{fig:umc_sample}).
}

\subsection{Supplementary for Dataset}
\subsubsection{Attribute Selection} 
\label{sec:att_selection}
\begin{figure*}[ht!]
    \centering
    \includegraphics[width=1\textwidth]{attribute_selection.pdf}
    \caption{
    \walon{Colorized cells in this table present the indexes of 239 CUB attributes used in our experiments (i.e. $\mathbf{A}^s \cap \mathbf{A}^u$), in which their corresponding base attributes are indicated in the black-shaded cells (i.e. $\mathbf{B}^c$ on the left-most column while $\mathbf{B}^p$ on the top row). For instance, the $279^{th}$ attribute in CUB is ``blue beak'', so we put ``279'' in the cell where its horizontal position in the table coincides with the one of the base attribute ``beak'', and its vertical position in table coincides with the one of the base attribute ``blue''.
    Cells with the same background color are in the same group.}}
    \label{fig:att_selection}
\end{figure*}

\walon{
As previously stated in our main manuscript, the CUB dataset has 312 attributes in total, each of which could be decomposed into an adjective and an object part.~(e.g., ``solid'' and ``breast'' for attribute ``solid breast'';  ``red'' and ``throat'' for attribute ``red throat''). The meanings behind the adjectives contain color, texture, shape, and others, while color (to which 239 of 312 attributes are related) is the dominant one. We thus focus on these 239 attributes (which have adjectives for color) in CUB and construct a table summarizing their corresponding base attributes (in total, 16 base attributes of object parts and 15 base attributes of colors)
as shown in Figure \ref{fig:att_selection} (please check the caption for interpreting this table). 
Please note that, though ideally there should be 240 attributes produced by all the combinations from 16 base attributes of object parts and 15 base attributes of colors, we do not have the attribute ``iridescent eye'' as it has no example shown in the CUB dataset. Therefore, the number of attributes used in our experiments is one less 240 (i.e., 239 attributes in total).}

\walon{
We divide the 239 attributes into 15 groups such that each of them has all the base attributes (i.e., 16 for object parts and 15 for colors) included (except for group 10, owing to the absent attribute: ``iridescent eye''). The attributes assigned to each of these 15 groups can be found in Figure \ref{fig:att_selection} (grouped by the cells with different color backgrounds). Such grouping helps us select the minimum number of seen attributes required for learning to synthesize the novel ones in a more efficient way, as the attributes from any two different groups (excluding group10) can be used to factor out all the base attributes via our intersection function $\mathbb{I}$. Please note that there exist more than one possible ways of grouping to achieve the same goal; here, we only describe the way used in our experiments.}

\walon{In our experimental settings, we use group1 and group2 as seen attributes $\mathbf{A}^s$ for the experiments of $N^s=32$ (cf. Table.1 and Table.2 in our main manuscript). For the experiments of $N^s=64$, group1, group2, group3, and group4 are used as seen attributes. Moreover, for the experiments of $N^s=96$, group1 to group6 are used together as seen attributes. Next, we conducted a study to verify the consistency of our proposed method to different combinations of seen attributes. We randomly select two groups as seen attributes (i.e., $N^s=32$) to train our decompose-and-reassemble procedure and evaluate the performance of synthesized novel attribute detectors. In total, we repeat this experiment for six rounds. The standard deviations of three metrics (i.e., mAUROC, mAP@50, and mLA) among these 6 rounds are $0.0056$, $0.0124$, and $0.0175$, respectively. The relatively low variance thus successfully verifies the consistency of our proposed method to various combinations of seen attributes.}

\subsubsection{Details of $\alpha$-CLEVR}
\label{appendix:alpha clevr}

\begin{figure*}[ht!]
    \centering
    \includegraphics[width=1\textwidth]{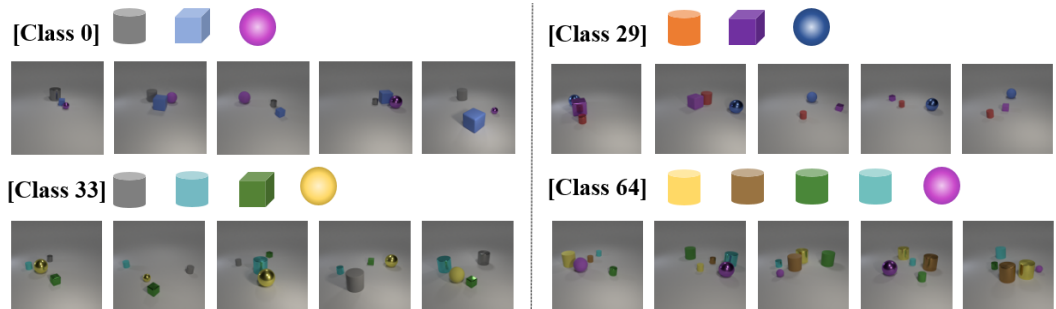}
    \vspace{-2em}
    \caption{\appendixNew{Samples from our $\alpha$-CLEVR dataset. Classes in $\alpha$-CLEVR dataset are defined by the specific combinations of toy bricks (where toy bricks with different color-shape combinations are treated as different attributes). Note that the images of the same class would have variances in terms of material, size, and relative locations of the toy bricks.
    }}
    \label{fig:ClevrSamples}
    \vspace{-1em}
\end{figure*}
\begin{figure}[ht!]
    \centering
    \includegraphics[width=0.40\textwidth]{attribute_selection_clevr.pdf}
    \vspace{-1em}
    \caption{
    \appendixNew{Colorized cells in this table present the indexes of 24 attributes used in our $\alpha$-CLEVR experiments (i.e. $\mathbf{A}^s \cap \mathbf{A}^u$), in which their corresponding base attributes are indicated in the black-shaded cells (i.e. $\mathbf{B}^c$ on the left-most column while $\mathbf{B}^p$ on the top row). For instance, the first attribute in $\alpha$-CLEVR is ``gray cube'', so we put ``1'' in the cell where its horizontal position in table coincides with the one of the base attribute ``cube'', and its vertical position in the table coincides with the one of the base attribute ``gray''.
    Cells with the same background color are in the same group.}}
    \label{fig:att_selection_clevr}
    \vspace{-1em}
\end{figure}
\begin{figure}[ht!]
    \centering
    \includegraphics[width=0.45\textwidth]{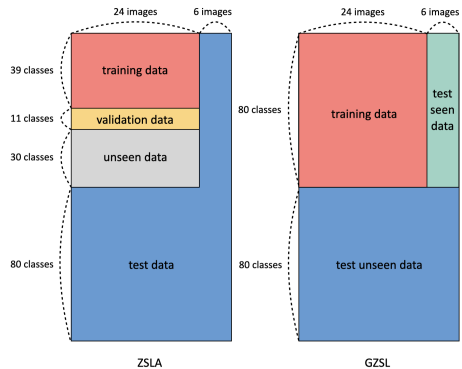}
    \vspace{-1em}
    \caption{
    \appendixNew{Train/test split of ZSLA (left part) and GZSL (right part): For the scenario of ZSLA, we will first use the training data to obtain seen attribute detectors and then use decompose-and-reassemble algorithm to synthesize unseen attribute detectors; For the scenario of GZSL, we annotate/re-annotate the dataset via attribute detectors synthesized by ZSLA and use them to compute the class-attribute matrix for GZSL training. Finally, we evaluate the quality of our attribute detectors by the test data as shown in this figure~(i.e. blue area in the left part, which is the same as blue and green area in the right part).}}
    \label{fig:split}
\end{figure}

\appendixNew{
$\alpha$-CLEVR dataset is a modification of~\cite{clevr}, in which~\cite{clevr} not only offers a well-known diagnostic dataset:~``CLEVR'' for VQA tasks but also provides a framework for people to create their dataset with different purposes. The official CLEVR dataset contains 100,000 images composed of several toy bricks. Eight colors and three shapes are used to describe these bricks. Due to the missing concept of \textbf{class} in the official CLEVR dataset, we define ours based on the released program and name our dataset $\alpha$-CLEVR.
}

\appendixNew{
In detail, we adopt colors and shapes as the \textbf{base attribute set} and treat the color-shape combinations for bricks as the \textbf{attribute set} (i.e., in total there are 24 attributes, representing red cube, blue sphere, etc.). Figure~\ref{fig:att_selection_clevr} shows the base attributes and their combinations (in the same way as CUB shown in Figure~\ref{fig:att_selection}). The 24 attributes in $\alpha$-CLEVR dataset are divided into three groups. Each of them contains all of the base attributes. The grouping method is under the same scheme as what we used in the CUB dataset to effectively utilize the seen attributes~(group1 and group2 in our experimental setting).
On the other hand, a \textbf{class} can be defined as a specific combination of attributes (e.g. an image having gray cylinder, blue cube, and purple sphere is belonging to the class ``GrayCylinder-BlueCube-PurpleSphere''). Furthermore, since real-world datasets usually would contain many non-class-related factors, such as items that appear in different poses or color variance caused by different cameras, we hence introduce several factors of variance (such as the relative location, materials, and the size of the bricks) into our $\alpha$-CLEVR to mimic the real-world scenario. We show some image examples of our $\alpha$-CLEVR dataset in Figure~\ref{fig:ClevrSamples}, where the images from the same class have the same combination of toy bricks (i.e. the same color-shape attributes) but would have variances in terms of materials, sizes, and relative locations between toy bricks.
}

\appendixNew{Figure~\ref{fig:split} shows the train/test split on our $\alpha$-CLEVR dataset for the two scenarios: learning our proposed ZSLA task (i.e. Zero-Shot Learning for Attributes) or learning GZSL (i.e. Generalized Zero-Shot Learning). As mentioned in Section~4
~of the main manuscript, each class has 30 images; 80 classes are used for GZSL training on $\alpha$-CLEVR dataset, and the other disjoint 80 classes are set as unseen test data. Among the 80 seen classes, 50 classes (39 classes for training, 11 classes for validation) composed of seen attributes $\mathbf{A}^s$ are used to synthesize unseen attribute detectors in ZSLA, and the other 30 classes containing novel attributes $\mathbf{A}^u$ will be isolated from ZSLA training. We use the unseen attribute detectors together with the seen ones to annotate all attribute labels in the $\alpha$-CLEVR dataset and obtain the class-wise statistics of attribute labels to form the class-attribute matrix (i.e., the semantic information for classes).
Note that we even use our seen attribute detectors to re-annotate the attributes of the training images in Section 4.3 of the main manuscript
~due to their noisy attribute labels. After that, the annotated dataset with the class-attribute matrix can be further utilized by GZSL algorithms.
During the evaluation period, we use the same test data to measure the quality of our unseen attribute detectors via: (1) mAUROC, mAP@50, and mLA of novel attribute annotations to test for attribute classification, attribute retrieval, and attribute localization respectively; (2) the performance of GZSL trained with the attribute labels which are re-annotated by the detectors obtained from ZSLA.
}
\subsection{Supplementary for Experiments}
\subsubsection{Further Discussion on Experimental Results of Automatic Annotations for Learning Generalized Zero-Shot Image Classification}
\label{appendix:label consistency problem}

Furthermore, we give a deeper discussion on why the annotations provided by our synthesized attribute detectors can improve the GZSL performance compared with the results upon manual annotations (cf. Table~\ref{tab:deltaCUB_GZSL} in Section ~\ref{sec:reannotate}). 

\begin{figure*}[ht!]
    \centering
    \includegraphics[width=.7\textwidth]{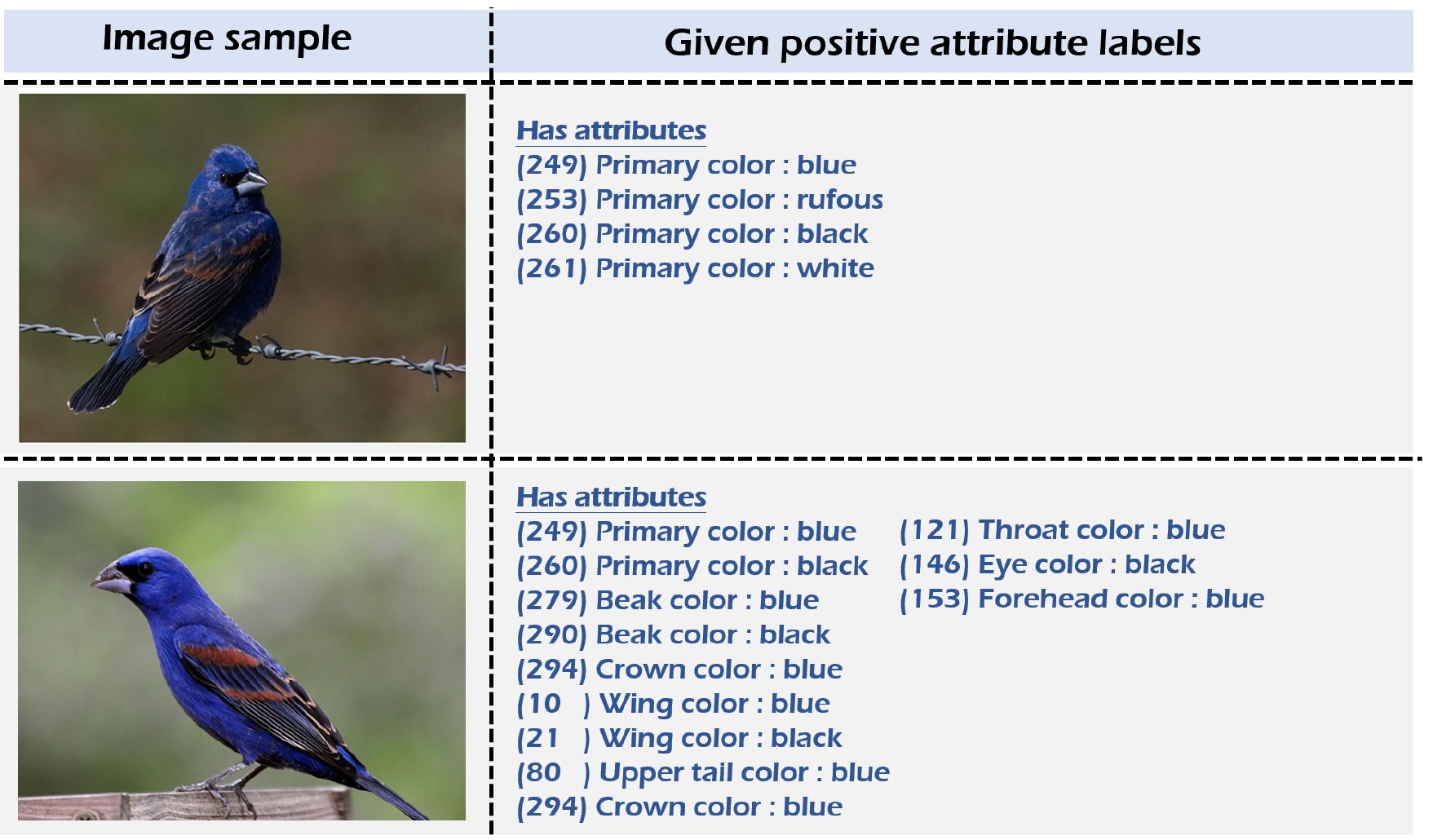}
    \caption{An example from the CUB dataset demonstrates the issue of attribute label inconsistency across the bird images of the same species. The number before each attribute description is the corresponding attribute index defined in Fig~\ref{fig:att_selection}.}
    \label{fig:manual_label}
\end{figure*}

\walon{As mentioned in Section~\ref{sec:reannotate} 
of our main manuscript, the inconsistency between different human annotators when building the CUB dataset would cause noisy/ambiguous attribute labels. Figure~\ref{fig:manual_label} shows an example with such ambiguity/noise, where two bird images of the same species sharing similar visual appearance are manually annotated with quite different attribute labels. For the upper image, the annotator may treat the crown, beak, and others as a whole to be the primary body, thus only the adjective descriptions of the primary body part are labeled. The second annotator distinguishes different parts and gives precise and more fine-grained part descriptions for the bottom image. Such label inconsistency across images is harmful to model learning. In this example, the confusing label ``primary blue" (i.e., instead of the precise label ``blue crown") would introduce unnecessary biases into the class-attribute matrix and hence have a negative impact on the final performance for the GZSL task.}

\walon{On the other hand, the machine-annotated $\delta$-CUB dataset created by our synthesized attribute detectors can mitigate this inconsistency issue from two aspects. First, the machine-annotated dataset is labeled based on a unified model instead of multiple annotators and hence can somehow avoid the issue of label inconsistency. Second, although our model learns the seen attribute detectors from noisy human attribute annotation, the extracted attribute classifiers could be more robust to the inconsistency of attribute labels due to the usage of many training images as well as the location information for training. By extracting the representative detectors from many images of the same attribute (even some labels might be noisy), the influence from inconsistent labels can be implicitly reduced. Also, the location information, forcing the representative detectors to highlight the target parts accurately, can significantly clarify the ambiguous part labels introduced by annotators (e.g., the primary and crown example mentioned before). Thus, the synthesized detectors, which are learnt by our proposed method from a set of seen attribute detectors (that are less sensitive to inconsistent labels) are able to provide more robust machine annotations.
}  
\subsubsection{Details of Obtaining Class-attribute Matrix for $\delta$-CUB}
Here we give a detailed discussion on how we generate the class-attribute matrix for~$\delta$-CUB. The class-attribute matrix plays an essential role in the zero-shot classification task to associate the categories by describing them as the composition of attributes. The meaning of each entry in the class-attribute matrix (in size of ``number of categories'' $\times$ ``number of attributes'') can be roughly understood as "what percentage of instances in a category are considered to have a certain attribute". In the CUB dataset, to build the class-attribute matrix, they random sample some images from a category and ask multiple workers to annotate these images several times, and then the percentage of assigning different attributes to the images will be treated as the attribute composition of this category. 
\appendixNew{As our proposed method is able to automatically annotate instance-level attribute labels, in order to mimic the way CUB works, we binarize the posterior probability of detecting an attribute given a test image (i.e. $\sigma(\tilde{r}_k(x))$ as Equation.\ref{eq:seen_ce_no_location} in our main manuscript, indicating the posterior probability of having the $k$-th attribute in image $x$). Regarding the threshold to binarize the posterior, it is determined by maximizing $TPR-FPR$ over all the seen attributes, where $TPR$ and $FPR$ are the true positive rate and the false positive rate respectively.}

\subsubsection{Baselines}
\label{appendix:baseline}
\modify{
As stated in Section~\ref{sec:exp_result}
, existing GZSL algorithms cannot do ZSL on attributes directly.}
\modify{Hence, to fit our proposed problem scenario, there are several modifications on their original formulation to achieve the adaption: (1) replacing class/attribute with attribute/base-attribute,
(2) changing the task setting from multi-class to multi-attribute binary classification, and (3) switching image-wise feature representations to patch-wise ones.}

\modify{
Note that the three modifications we list are not simple for every GZSL algorithm. 
For example, some embedding-based methods like~\cite{ALE, SJE} learn the mapping function from image feature space to latent space defined by side information. Such approaches classify the categories based on the distance to each prototype, which makes it not easy to turn the task setting from multi-class to multi-attribute binary classification.
Furthermore, generative-based GZSL methods like~\cite{f-WGAN, CADAVAE, f-VAEGAN-D2} are much more complicated and the hyper-parameters in each approach must be re-tuned to fit the new task.}

\modify{On the contrary, the idea of ESZSL~\cite{ESZSL} and LAGO~\cite{LAGO} based on the implicit assumption of logical operation on multiple classifiers are more suitable to be treated as the baseline methods. Note that we actually implement LAGO-Singleton mentioned in~\cite{LAGO}, which could be viewed as the relaxed version of DAP~\cite{DAP_IAP}.}
\subsubsection{Additional Results of $\alpha$-CLEVR}
\begin{figure*}[h!]
    \centering
    \hspace{0.5em}
    \includegraphics[width=0.9\textwidth]{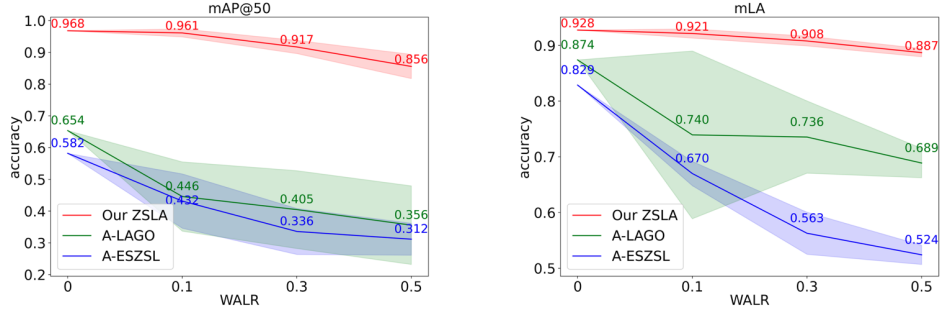}
    \vspace{-1em}
    \caption{\appendixNew{
    Evaluation (in terms of attribute retrieval and attribute localization, with mAP@50 and mLA as metrics respectively) on the robustness against noisy attribute labels for various methods which learn to synthesize the novel attribute detectors. The shaded bands around each curve represent the 95\% confidence interval over 5 runs of different noisy label sets.
    }}
    \label{Fig.CI_mAP_at_50_mLA}
\end{figure*}
\appendixNew{As described in Section~\ref{sec:robustness} 
of the main manuscript, we show the robustness of ZSLA against the noisy labels 
in terms of mAUROC. In addition to attribute classification, we also adopt attribute retrieval and attribute localization~(with mAP@50 and mLA as metrics, respectively) to further demonstrate the robustness of ZSLA. As the mAP@50 and mLA plots provided in Figure~\ref{Fig.CI_mAP_at_50_mLA}, we observe that: (1) our ZSLA surpasses baselines in attribute retrieval and localization for all the \textbf{WALRs}; (2) our ZSLA is more robust than baselines as indicated by having less performance drop when \textbf{WALR} is increased; (3) in comparison to baselines, our ZSLA has a lower variance over multiple runs of different noisy label sets. All three statements coincide with our observations in Section~\ref{sec:robustness} 
of the main manuscript and further verify the robustness of our proposed ZSLA.
}
\subsection{Extensive Experiments}

\subsubsection{Ablation Study}
\label{appendix:ablation study}
\setlength{\tabcolsep}{0.9mm}
\begin{table}[h!]
\centering
\caption{
\walon{
Quantitative evaluation (in terms of attribute classification, retrieval, and localization) on the novel attribute detectors learnt by three model variants, in order to have ablation study on the usages of uni-modal constraint (abbreviated as ``UMC'', implemented by $\mathcal{L}_{umc}$) and location information (abbreviated as ``Loc Info'').}}
\begin{tabular}{cc|ccc}
\textbf{Loc Info} & \textbf{UMC} & \textbf{mAUROC} & \textbf{mAP@50} & {\color[HTML]{333333} \textbf{mLA}} \\ \hline\hline
\ding{51} & \ding{51}  & .689 & .320 & .846 \\
\ding{55}  & \ding{51}  & .701 & .296 & .613 \\
\ding{55} & \ding{55} & .702 & .325 & .348
\end{tabular}
\label{tab:ablation_study}
\end{table}

\walon{
Here, we conduct an ablation study and investigate the influence/impact of \textbf{1)} the ``\textbf{uni-modal constraint}'' (abbreviated as UMC, implemented by $\mathcal{L}_{umc}$ in our proposed method, cf. Equation
of our main manuscript), and \textbf{2)} the usage of the ground-truth of the attribute locations (i.e. knowing where an attribute appears on the image, denoted as ``\textbf{location information}'') in training the seen attribute detectors. 
\modify{Note that $\mathcal{L}_{bce}$ in Equation~\ref{eq:seen_ce_no_location} 
and $\mathcal{L}_{rec}$ in Equation~\ref{eq:rec} 
are the fundamental objective functions to train the seen attribute detectors and the intersection model $\mathbb{I}$, respectively; thus, these two functions are must-have components in our proposed ZSLA approach and are excluded from the ablation study.}
Ideally, we expect that: if the seen attribute detectors are better trained, it is more likely to obtain the synthesized attribute detectors with better performance (as those seen attribute detectors are the input materials for learning decompose-and-reassemble procedure).}

\walon{
The evaluation results on the synthesized novel attributes learnt by adopting different usage combinations of the uni-modal constraint and the location information are summarized in Table~\ref{tab:ablation_study}. We are able to observe that: (1)
With the help of uni-modal constraint, the mLA (i.e. average localization accuracy) of synthesized novel attributes clearly improves (i.e. from 0.348 to 0.613); (2) In addition to the uni-modal constraint, if the location information is also considered during the model training, the mLA can even go further to gain an extra boost by 0.233 (i.e. from 0.613 to 0.846). The overall improvements in terms of mLA made by having both uni-modal constraint and location information adopted in training our proposed method clearly indicate their effectiveness to help precisely extract and synthesize novel attributes. 
}

\begin{figure*}[t] 
\centering 
\includegraphics[width=1\textwidth]{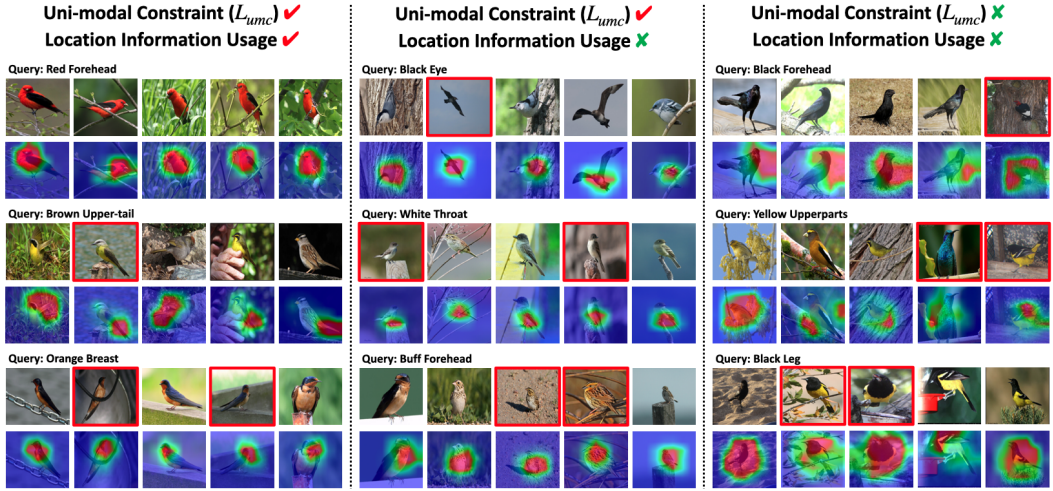} 
\vspace{-1.5em}
\caption{
\walon{Example results of attribute retrieval and localization for the novel attribute detectors learnt by three model variants, in order to have ablation study on the usages of uni-modal constraint (abbreviated as UMC, implemented by $\mathcal{L}_{umc}$) and location information. These three model variants are trained (left) with UMC and location information, (middle) with UMC but without location information, and (right) with neither UMC nor location information. For each example set, we show the top-5 retrieved images and their response maps for a synthesized novel attribute, where the images marked with red borders are the false positives according to CUB ground-truth.}
}
\label{Fig.ablation_retrieval} 
\end{figure*}

\walon{
This study also finds that: as both mAUROC and mAP@50 metrics (which are related to attribute classification and retrieval) do not aim to localize the image regions of the target attributes, they are hence relatively insensitive to the usage of uni-modal constraint and location information.
Some qualitative examples of this ablation study are provided in Figure~\ref{Fig.ablation_retrieval}. We can see that: Without using uni-modal constraint and location information (cf. the right portion of Figure~\ref{Fig.ablation_retrieval}), the response maps of the target novel attributes show multiple modes on wrong locations; after introducing the uni-modal constraint, the response maps turn to have more concentrated distribution (i.e. uni-modal) but occasionally have the modes on the incorrect locations for the target attributes (cf. the middle portion of Figure~\ref{Fig.ablation_retrieval}); upon further taking the location information into consideration for model training, the localization of the target attribution is improved and becomes more accurate (cf. the left portion of Figure~\ref{Fig.ablation_retrieval}).
}

\subsubsection{The Usage of Location Information}\label{sec:loc_info}
\modify{In our previous experiments, both our ZSLA and baselines are trained with the auxiliary location information, which might not be available in other dataset~(or might be viewed as an extra cost for dataset establishment). Besides, from the ablation study that we discussed previously, we observe that the additional location information mainly benefits attribute localization ability while having a relatively minor impact on attribute retrieval and classification, which are the crucial metrics for attribute annotations.
Consequently, we replicate our experiments of Section~\ref{sec:experiment1} 
and Section~\ref{sec:reannotate} 
of the main manuscript in the same setting except that we do not utilize the auxiliary location information provided by CUB dataset.}
\begin{center}
\setlength{\tabcolsep}{2mm}
\begin{table}[h]
\caption{\modify{Without the usage of location information during model training, the evaluation of synthesized novel/unseen attributes on attribute classification (mAUROC), retrieval (mAP@50), and localization (mLA). $N^s$ is the number of seen attributes.}
}
\centering
\begin{tabular}{c|c|ccc}
 &
  $N^s$ &
  mAUROC &
  mAP@50 &
  mLA \\ \hline\hline
 &
  {\color[HTML]{3531FF} 32} &
  {\color[HTML]{3531FF} .565} &
  {\color[HTML]{3531FF} .235} &
  {\color[HTML]{3531FF} .184} \\
 &
  {\color[HTML]{009901} 64} &
  {\color[HTML]{009901} .608} &
  {\color[HTML]{009901} .277} &
  {\color[HTML]{009901} .265} \\
\multirow{-3}{*}{\textbf{A-ESZSL}} &
  {\color[HTML]{963400} 96} &
  {\color[HTML]{963400} .638} &
  {\color[HTML]{963400} .294} &
  {\color[HTML]{963400} .269} \\ \hline
 &
  {\color[HTML]{3531FF} 32} &
  {\color[HTML]{3531FF} .608} &
  {\color[HTML]{3531FF} .231} &
  {\color[HTML]{3531FF} .324} \\
 &
  {\color[HTML]{009901} 64} &
  {\color[HTML]{009901} .633} &
  {\color[HTML]{009901} .260} &
  {\color[HTML]{009901} .371} \\
\multirow{-3}{*}{\textbf{A-LAGO}} &
  {\color[HTML]{963400} 96} &
  {\color[HTML]{963400} .654} &
  {\color[HTML]{963400} .284} &
  {\color[HTML]{963400} .389} \\ \hline
 &
  {\color[HTML]{3531FF} \textbf{32}} &
  {\color[HTML]{3531FF} \textbf{.700}} &
  {\color[HTML]{3531FF} \textbf{.296}} &
  {\color[HTML]{3531FF} \textbf{.613}} \\
 &
  {\color[HTML]{009901} \textbf{64}} &
  {\color[HTML]{009901} \textbf{.720}} &
  {\color[HTML]{009901} \textbf{.332}} &
  {\color[HTML]{009901} \textbf{.662}} \\
\multirow{-3}{*}{\textbf{Our ZSLA}} &
  {\color[HTML]{963400} \textbf{96}} &
  {\color[HTML]{963400} \textbf{.738}} &
  {\color[HTML]{963400} \textbf{.339}} &
  {\color[HTML]{963400} \textbf{.662}}
\end{tabular}
\label{tab:table_no_loc}
\vspace{-5mm}
\end{table}
\end{center}

\begin{table*}[t]
\vspace{1em}
\caption{Experiments results to replicate the ones in Section~4.2 
of our main manuscript but particularly without using the auxiliary location information provided by CUB dataset.}

\resizebox{\textwidth}{!}{
\begin{tabular}{c|cccc|cccc|cccc|cccc|}
\cline{2-17}
 &
  \multicolumn{4}{c|}{CADAVAE~\cite{CADAVAE}} &
  \multicolumn{4}{c|}{TFVAEGAN~\cite{tfVAEGAN}} &
  \multicolumn{4}{c|}{ALE~\cite{ALE}} &
  \multicolumn{4}{c|}{ESZSL~\cite{ESZSL}} \\ \cline{2-17} 
 &
  S &
  U &
  H &
  {\color[HTML]{B234FC} \textbf{GAIN}} &
  S &
  U &
  H &
  {\color[HTML]{B234FC} \textbf{GAIN}} &
  S &
  U &
  H &
  {\color[HTML]{B234FC} \textbf{GAIN}} &
  S &
  U &
  H &
  {\color[HTML]{B234FC} \textbf{GAIN}} \\ \hline
\rowcolor[HTML]{ECF4FF} 
\multicolumn{1}{|c|}{\cellcolor[HTML]{ECF4FF}\begin{tabular}[c]{@{}c@{}}Manual\\ ($N^s$=32 for CUB)\end{tabular}} &
  42.9 &
  27.3 &
  33.4 &
  - &
  45.5 &
  31.2 &
  37.1 &
  - &
  26.4 &
  9.2 &
  13.7 &
  - &
  29.8 &
  10.8 &
  15.9 &
  - \\
\rowcolor[HTML]{E6FFE6} 
\multicolumn{1}{|c|}{\cellcolor[HTML]{E6FFE6}\begin{tabular}[c]{@{}c@{}}Manual\\ ($N^s$=312 for CUB)\end{tabular}} &
  {\color[HTML]{FE0000} \textbf{53.5}} &
  {\color[HTML]{3531FF} \textbf{51.6}} &
  {\color[HTML]{3531FF} \textbf{52.4}} &
  {\color[HTML]{B234FC} +19.0} &
  {\color[HTML]{FE0000} \textbf{64.7}} &
  {\color[HTML]{3531FF} \textbf{52.8}} &
  {\color[HTML]{3531FF} \textbf{58.1}} &
  {\color[HTML]{B234FC} +21.0} &
  {\color[HTML]{FE0000} \textbf{62.8}} &
  {\color[HTML]{3531FF} \textbf{23.7}} &
  {\color[HTML]{3531FF} \textbf{34.4}} &
  {\color[HTML]{B234FC} +20.7} &
  {\color[HTML]{3531FF} \textbf{63.8}} &
  {\color[HTML]{3531FF} \textbf{12.6}} &
  {\color[HTML]{3531FF} \textbf{21.0}} &
  {\color[HTML]{B234FC} +5.1} \\
\rowcolor[HTML]{FFF3E4} 
\multicolumn{1}{|c|}{\cellcolor[HTML]{FFF3E4}A-LAGO} &
  {\color[HTML]{3531FF} \textbf{50.5}} &
  48.4 &
  49.4 &
  {\color[HTML]{B234FC} +16.0} &
  {\color[HTML]{3531FF} \textbf{62.5}} &
  48.0 &
  54.3 &
  {\color[HTML]{B234FC} +17.2} &
  44.0 &
  19.7 &
  27.3 &
  {\color[HTML]{B234FC} +13.6} &
  55.5 &
  12.2 &
  20.0 &
  {\color[HTML]{B234FC} +4.1} \\
\rowcolor[HTML]{FFF3E4} 
\multicolumn{1}{|c|}{\cellcolor[HTML]{FFF3E4}A-ESZSL} &
  46.9 &
  49.9 &
  48.4 &
  {\color[HTML]{B234FC} +15.0} &
  60.6 &
  49.2 &
  54.3 &
  {\color[HTML]{B234FC} +17.2} &
  44.3 &
  22.2 &
  29.6 &
  {\color[HTML]{B234FC} +15.9} &
  {\color[HTML]{FE0000} \textbf{64.6}} &
  7.5 &
  13.4 &
  {\color[HTML]{B234FC} -2.5} \\
\rowcolor[HTML]{FFF3E4} 
\multicolumn{1}{|c|}{\cellcolor[HTML]{FFF3E4}\begin{tabular}[c]{@{}c@{}}Our ZSLA\\ ($N^s$=32,  $N^u$=207 for $\delta'$-CUB)\end{tabular}} &
  {\color[HTML]{333333} 50.0} &
  {\color[HTML]{FE0000} \textbf{57.7}} &
  {\color[HTML]{FE0000} \textbf{53.6}} &
  {\color[HTML]{B234FC} {\ul \textbf{+20.2}}} &
  61.9 &
  {\color[HTML]{FE0000} \textbf{58.4}} &
  {\color[HTML]{FE0000} \textbf{60.1}} &
  {\color[HTML]{B234FC} {\ul \textbf{+23.0}}} &
  {\color[HTML]{3531FF} \textbf{52.2}} &
  {\color[HTML]{FE0000} \textbf{32.0}} &
  {\color[HTML]{FE0000} \textbf{39.7}} &
  {\color[HTML]{B234FC} {\ul \textbf{+26.0}}} &
  55.7 &
  {\color[HTML]{FE0000} \textbf{22.4}} &
  {\color[HTML]{FE0000} \textbf{32.0}} &
  {\color[HTML]{B234FC} {\ul \textbf{+16.1}}} \\ \hline
\end{tabular}
}
\label{tab:table_no_loc_GZSL}
\end{table*}

\modify{Table~\ref{tab:table_no_loc} shows the performance in terms of mAUROC, mAP@50, mLA of ZSLA, and baselines with $N^s$ set as ${32, 64, 96}$. Similar results are observed as what we have in Section~\ref{sec:experiment1} 
of the main manuscript, our ZSLA outperforms baselines on all $N^s$ settings and evaluation schemes, verifying the superior ability of our ZSLA in terms of attribute classification, retrieval, and localization in comparison with baselines no matter if the location information is accessible or not.}  

\modify{Afterwards, we utilize the 32 seen attribute detectors and 207 synthesized novel attribute detectors to re-annotate CUB dataset as $\delta'$-CUB.~(to be differentiated from $\delta$-CUB, which adopts location information during the training of seen attribute detectors); in parallel, baselines are also used to re-annotate CUB dataset. Then GZSL algorithms~(i.e. CADAVAE, TFVAEGAN, ALE, ESZSL) are trained and evaluated by $\delta'$-CUB and the results are summarized in the row shaded by the orange color of Table~\ref{tab:table_no_loc_GZSL}. Apart from the orange part, we report the experimental results on training the four GZSL methods by using only 32 attributes or using all 312 attributes provided by the original CUB again and summarize them in the rows respectively shaded by the blue and green color of Table~\ref{tab:table_no_loc_GZSL}. From the results, we can clearly see that GZSL algorithms make significant improvement when using either baselines or our ZSLA to automatically annotate all the attributes in $\delta'$-CUB and our ZSLA has relatively large gains on the performance of GZSL algorithms in comparison to the baselines. This experiment setting makes sure that the rows shaded by the orange color are exactly at the same cost (in terms of the human efforts to annotate the seen attributes) as the rows shaded by the blue color, which further verify the usefulness of our ZSLA (i.e. improving the performance without any additional cost on human annotations).
}

\subsubsection{Comparison to CZSL}
\label{appendix:CZSL}

\nips{As stated in Section 2 of our main manuscript, \textit{compositional zero-shot learning} (CZSL) is  conceptually related to (but different from) our task scenario. Here, we would like to again emphasize that there exists significant differences between our proposed scenario and the existing CZSL setting: (1) An image in our problem scenario would have multiple attributes while there usually exists only a single state-object composition for CZSL; (2) The attribute detectors synthesized by our proposed ZSLA are able to automatically provide the labels of novel attributes (i.e. these novel attributes do not have any manually labeled samples in the training set) for all the images thus leading to more detailed descriptions for all the categories, while CZSL typically aims to increase the number of categories. Furthermore, many of the CZSL algorithms~\cite{nagarajan2018attributes, SymNet, TMN, huynh2020compositional, naeem2021learning} proposed utilizing auxiliary information such as word2vec~\cite{W2V} or GloVe~\cite{GloVe} to get a better initialization for the embedding, which raises the additional cost.
On the other hand, RedWine~\cite{misra2017red} (a representative approach of CZSL) proposed to train a set of SVM and take the weight parameters as the initialized embedding for CZSL. However, training the SVMs for the base attribute can be suffered from the extreme label distribution of the base attributes. In our problem scenario, an image usually contains multiple attributes which are the combinations of the shared set of base attributes; in other words, it is common to see some base attributes to appear in almost every image. We demonstrate the extreme label distribution of the base attribute in CUB in the second and third columns of Table~\ref{tab:bacc_distribution}. What's worse, the base attributes with their positive labels in a high ratio might even co-occur in the training images frequently. These challenges obstruct the SVM to capture correct features for each base attribute and thus make the SVM fail to be representative. Comparing the AUROC of every single SVM on the training set and the testing set in the fourth and fifth column of Table~\ref{tab:bacc_distribution}, it is clear that the learned SVMs are not generalized at all.}

\begin{table}[h!]
\caption{\modify{The table demonstrates the extreme label distribution of base attributes in the CUB dataset as well as the performance of SVMs trained upon the labels\walon{, which would cause problems for the representative CZSL baseline RedWine~\cite{misra2017red}}. During training the compositional network, the embedding of the base attribute ``Primary" is set as a random weight due to the fact that it appears in every single image, and thus it is impossible to train an SVM for ``Primary".}}
\resizebox{\textwidth}{!}{
\begin{tabular}{|c|cc|cc|lccccc}
\cline{1-5} \cline{7-11}
\textbf{}                                                                      & \multicolumn{2}{c|}{\textbf{Positive label   ratio (\%)}}                        & \multicolumn{2}{c|}{\textbf{AUROC}}                                          & \multicolumn{1}{l|}{} & \multicolumn{1}{c|}{\textbf{}}                                                                      & \multicolumn{2}{c|}{\textbf{Positive label   ratio (\%)}}                              & \multicolumn{2}{c|}{\textbf{AUROC}}                                            \\ \cline{1-5} \cline{7-11} 
\textbf{\begin{tabular}[c]{@{}c@{}}Name of the \\ base attribute\end{tabular}} & \multicolumn{1}{c|}{\textbf{Training set}}        & \textbf{Testing set}         & \multicolumn{1}{c|}{\textbf{Training}}          & \textbf{Testing}           & \multicolumn{1}{l|}{} & \multicolumn{1}{c|}{\textbf{\begin{tabular}[c]{@{}c@{}}Name of the \\ base attribute\end{tabular}}} & \multicolumn{1}{c|}{\textbf{Training set}} & \multicolumn{1}{c|}{\textbf{Testing set}} & \multicolumn{1}{c|}{\textbf{Training}} & \multicolumn{1}{c|}{\textbf{Testing}} \\ \cline{1-5} \cline{7-11} 
\textbf{Beak}                                                                  & \multicolumn{1}{c|}{98.7}                         & 98.7                         & \multicolumn{1}{c|}{1.000}                      & 0.52                       & \multicolumn{1}{l|}{} & \multicolumn{1}{c|}{\textbf{Blue}}                                                                  & \multicolumn{1}{c|}{10.3}                  & \multicolumn{1}{c|}{11.0}                 & \multicolumn{1}{c|}{0.515}             & \multicolumn{1}{c|}{0.497}            \\ \cline{1-5} \cline{7-11} 
\textbf{Wing}                                                                  & \multicolumn{1}{c|}{94.7}                         & 93.1                         & \multicolumn{1}{c|}{0.999}                      & 0.581                      & \multicolumn{1}{l|}{} & \multicolumn{1}{c|}{\textbf{Brown}}                                                                 & \multicolumn{1}{c|}{43.8}                  & \multicolumn{1}{c|}{42.6}                 & \multicolumn{1}{c|}{0.486}             & \multicolumn{1}{c|}{0.473}            \\ \cline{1-5} \cline{7-11} 
\textbf{Upper-parts}                                                           & \multicolumn{1}{c|}{87.4}                         & 85.9                         & \multicolumn{1}{c|}{0.963}                      & 0.63                       & \multicolumn{1}{l|}{} & \multicolumn{1}{c|}{\textbf{Iridescent}}                                                            & \multicolumn{1}{c|}{5.8}                   & \multicolumn{1}{c|}{8.5}                  & \multicolumn{1}{c|}{0.484}             & \multicolumn{1}{c|}{0.487}            \\ \cline{1-5} \cline{7-11} 
\textbf{Under-parts}                                                           & \multicolumn{1}{c|}{90.3}                         & 90.3                         & \multicolumn{1}{c|}{0.999}                      & 0.645                      & \multicolumn{1}{l|}{} & \multicolumn{1}{c|}{\textbf{Purple}}                                                                & \multicolumn{1}{c|}{2.1}                   & \multicolumn{1}{c|}{2.8}                  & \multicolumn{1}{c|}{0.457}             & \multicolumn{1}{c|}{0.482}            \\ \cline{1-5} \cline{7-11} 
\textbf{Breast}                                                                & \multicolumn{1}{c|}{92.1}                         & 92.2                         & \multicolumn{1}{c|}{0.998}                      & 0.574                      & \multicolumn{1}{l|}{} & \multicolumn{1}{c|}{\textbf{Rufous}}                                                                & \multicolumn{1}{c|}{7.5}                   & \multicolumn{1}{c|}{6.9}                  & \multicolumn{1}{c|}{0.516}             & \multicolumn{1}{c|}{0.484}            \\ \cline{1-5} \cline{7-11} 
\textbf{Back}                                                                  & \multicolumn{1}{c|}{80.6}                         & 80.8                         & \multicolumn{1}{c|}{0.923}                      & 0.634                      & \multicolumn{1}{l|}{} & \multicolumn{1}{c|}{\textbf{Grey}}                                                                  & \multicolumn{1}{c|}{67.6}                  & \multicolumn{1}{c|}{66.5}                 & \multicolumn{1}{c|}{0.501}             & \multicolumn{1}{c|}{0.500}            \\ \cline{1-5} \cline{7-11} 
\textbf{Upper-tail}                                                            & \multicolumn{1}{c|}{67.1}                         & 65.0                         & \multicolumn{1}{c|}{0.675}                      & 0.54                       & \multicolumn{1}{l|}{} & \multicolumn{1}{c|}{\textbf{Yellow}}                                                                & \multicolumn{1}{c|}{27.1}                  & \multicolumn{1}{c|}{37.6}                 & \multicolumn{1}{c|}{0.421}             & \multicolumn{1}{c|}{0.413}            \\ \cline{1-5} \cline{7-11} 
\textbf{Throat}                                                                & \multicolumn{1}{c|}{95.2}                         & 95.2                         & \multicolumn{1}{c|}{0.551}                      & 0.508                      & \multicolumn{1}{l|}{} & \multicolumn{1}{c|}{\textbf{Olive}}                                                                 & \multicolumn{1}{c|}{6.5}                   & \multicolumn{1}{c|}{10.2}                 & \multicolumn{1}{c|}{0.420}             & \multicolumn{1}{c|}{0.443}            \\ \cline{1-5} \cline{7-11} 
\textbf{Eye}                                                                   & \multicolumn{1}{c|}{95.0}                         & 95.0                         & \multicolumn{1}{c|}{0.724}                      & 0.539                      & \multicolumn{1}{l|}{} & \multicolumn{1}{c|}{\textbf{Green}}                                                                 & \multicolumn{1}{c|}{3.8}                   & \multicolumn{1}{c|}{6.2}                  & \multicolumn{1}{c|}{0.534}             & \multicolumn{1}{c|}{0.468}            \\ \cline{1-5} \cline{7-11} 
\textbf{Forehead}                                                              & \multicolumn{1}{c|}{95.2}                         & 94.6                         & \multicolumn{1}{c|}{0.783}                      & 0.542                      & \multicolumn{1}{l|}{} & \multicolumn{1}{c|}{\textbf{Pink}}                                                                  & \multicolumn{1}{c|}{5.2}                   & \multicolumn{1}{c|}{3.8}                  & \multicolumn{1}{c|}{0.520}             & \multicolumn{1}{c|}{0.627}            \\ \cline{1-5} \cline{7-11} 
\textbf{Under\_tail}                                                           & \multicolumn{1}{c|}{74.6}                         & 74.1                         & \multicolumn{1}{c|}{0.523}                      & 0.479                      & \multicolumn{1}{l|}{} & \multicolumn{1}{c|}{\textbf{Orange}}                                                                & \multicolumn{1}{c|}{22.2}                  & \multicolumn{1}{c|}{18.8}                 & \multicolumn{1}{c|}{0.551}             & \multicolumn{1}{c|}{0.513}            \\ \cline{1-5} \cline{7-11} 
\textbf{Nape}                                                                  & \multicolumn{1}{c|}{92.8}                         & 91.9                         & \multicolumn{1}{c|}{0.706}                      & 0.548                      & \multicolumn{1}{l|}{} & \multicolumn{1}{c|}{\textbf{Black}}                                                                 & \multicolumn{1}{c|}{95.5}                  & \multicolumn{1}{c|}{95.7}                 & \multicolumn{1}{c|}{0.538}             & \multicolumn{1}{c|}{0.557}            \\ \cline{1-5} \cline{7-11} 
\textbf{Belly}                                                                 & \multicolumn{1}{c|}{89.2}                         & 89.6                         & \multicolumn{1}{c|}{0.68}                       & 0.616                      & \multicolumn{1}{l|}{} & \multicolumn{1}{c|}{\textbf{White}}                                                                 & \multicolumn{1}{c|}{64.4}                  & \multicolumn{1}{c|}{61.1}                 & \multicolumn{1}{c|}{0.457}             & \multicolumn{1}{c|}{0.452}            \\ \cline{1-5} \cline{7-11} 
{\color[HTML]{000000} \textbf{Primary}}                                        & \multicolumn{1}{c|}{{\color[HTML]{FF0000} 100.0}} & {\color[HTML]{FF0000} 100.0} & \multicolumn{1}{c|}{{\color[HTML]{FF0000} N/A}} & {\color[HTML]{FF0000} N/A} & \multicolumn{1}{l|}{} & \multicolumn{1}{c|}{\textbf{Red}}                                                                   & \multicolumn{1}{c|}{16.9}                  & \multicolumn{1}{c|}{13.7}                 & \multicolumn{1}{c|}{0.463}             & \multicolumn{1}{c|}{0.407}            \\ \cline{1-5} \cline{7-11} 
{\color[HTML]{000000} \textbf{Leg}}                                            & \multicolumn{1}{c|}{82.4}                         & 82.8                         & \multicolumn{1}{c|}{0.744}                      & 0.608                      & \multicolumn{1}{l|}{} & \multicolumn{1}{c|}{\textbf{Buff}}                                                                  & \multicolumn{1}{c|}{48.9}                  & \multicolumn{1}{c|}{46.7}                 & \multicolumn{1}{c|}{0.420}             & \multicolumn{1}{c|}{0.391}            \\ \cline{1-5} \cline{7-11} 
{\color[HTML]{000000} \textbf{Crown}}                                          & \multicolumn{1}{c|}{95.4}                         & 94.7                         & \multicolumn{1}{c|}{0.483}                      & 0.472                      &                       &                                                                                                     &                                            &                                           &                                        &                                       \\ \cline{1-5}
\end{tabular}
}
\label{tab:bacc_distribution}
\end{table}

\nips{\walon{Nevertheless, in order to demonstrate that our propose method does provide the state-of-the-art performance for the scenario of zero-shot learning on attributes, we try our best to adapt several CZSL algorithms into such scenario for making comparison, including}: RedWine~\cite{misra2017red}, LE+~\cite{misra2017red}~(which uses word2vec as the auxiliary information), AttOps~\cite{nagarajan2018attributes}, and CGE~\cite{naeem2021learning}. 
}
\nips{\walon{Particularly, in order to enable these algorithms to tackle our problem scenario of zero-shot learning on attributes}, 
\walon{we apply several modifications: \textbf{(1)} replacing state-object compositions (respectively, states or objects) with attributes (respectively, base attributes), and \textbf{(2)} changing their task setting from multi-class to multi-attribute binary classification (as in our scenario of zero-shot learning on attributes, an image could have multiple attributes), \textbf{(3)} switching image-wise feature representations to patch-wise ones, and \textbf{(4)} providing the additional location information during their training (for the purpose of making fair comparison as our method does use the location information).}
\walon{Moreover, regarding some specific objectives designed for predicting only a single state-object composition in an image (e.g. AttOps~\cite{nagarajan2018attributes} has several such objective functions), we respectively sample an attribute (i.e. ``attribute'' is analogous to ``state-object composition" in CZSL's scenario) from each image in the batch to calculate these objectives during each training iteration.
}
\walon{Other than the modifications described above, we keep their original hyperparameter settings as well as their original ways of initializing the embedding weights.}
\walon{Eventually, these CSZL algorithms after our adaptation/modification are able to tackle the unseen attributes and re-annotate the CUB dataset, hence we can make comparison with them (following the same evaluation schemes as shown in the main manuscript) for the scenario of zero-shot learning on attributes.}
}

\nips{\walon{The performance in terms of attribute classification, retrieval, and localization are summarized in Table~\ref{tab:CZSL_result1}. It is clear to see that, for all the evaluation metrics, our ZSLA consistently outperforms the CZSL algorithms.
The results for the re-annotation experiments (as conducted in Section~4.2 of the main manuscript) are provided in Table~\ref{tab:ZSLA-Base-GZSL}, where the performance of CZSL algorithms are shown in the purple-shaded rows. We can see that, the gain (with respect to the setting of using 32 manually-labeled attributes, i.e. the blue-shaded row) produced by our proposed ZSLA method (in the last orange-shaded row) is again significantly superior to the ones produced by the CZSL algorithms, even when some of the CZSL algorithms use the extra/auxiliary information such as word2vec. Such experimental results verify again the efficacy and the efficiency of our proposed ZSLA method. }}

\begin{center}
\setlength{\tabcolsep}{2mm}
\begin{table}[h!]
\centering
\caption{\nips{Evaluation of synthesized novel/unseen attributes on attribute classification (mAUROC), retrieval (mAP@50), and localization (mLA) on modified CZSL and our ZSLA. $N^s$ is the number of seen attributes.}
}
\begin{tabular}{c|c|ccc}
 & $N^s$ & mAUROC & mAP@50 & mLA \\ \hline\hline
 & {\color[HTML]{3531FF} 32} & {\color[HTML]{3531FF} .590} & {\color[HTML]{3531FF} .184} & {\color[HTML]{3531FF} .443} \\
 & {\color[HTML]{009901} 64} & {\color[HTML]{009901} .604} & {\color[HTML]{009901} .200} & {\color[HTML]{009901} .410} \\
\multirow{-3}{*}{\textbf{RedWine~\cite{misra2017red}}} & {\color[HTML]{963400} 96} & {\color[HTML]{963400} .622} & {\color[HTML]{963400} .240} & {\color[HTML]{963400} .415} \\ \hline
 & {\color[HTML]{3531FF} 32} & {\color[HTML]{3531FF} .596} & {\color[HTML]{3531FF} .210} & {\color[HTML]{3531FF} .499} \\
 & {\color[HTML]{009901} 64} & {\color[HTML]{009901} .615} & {\color[HTML]{009901} .240} & {\color[HTML]{009901} .485} \\
\multirow{-3}{*}{\textbf{LE+}~\cite{misra2017red}} & {\color[HTML]{963400} 96} & {\color[HTML]{963400} .638} & {\color[HTML]{963400} .260} & {\color[HTML]{963400} .485} \\ \hline
 & {\color[HTML]{3531FF} 32} & {\color[HTML]{3531FF} .630} & {\color[HTML]{3531FF} .253} & {\color[HTML]{3531FF} .382} \\
 & {\color[HTML]{009901} 64} & {\color[HTML]{009901} .640} & {\color[HTML]{009901} .293} & {\color[HTML]{009901} .380} \\
\multirow{-3}{*}{\textbf{AttOps}~\cite{nagarajan2018attributes}} & {\color[HTML]{963400} 96} & {\color[HTML]{963400} .670} & {\color[HTML]{963400} .322} & {\color[HTML]{963400} .490} \\ \hline
 & {\color[HTML]{3531FF} 32} & {\color[HTML]{3531FF} .601} & {\color[HTML]{3531FF} .266} & {\color[HTML]{3531FF} .363} \\
 & {\color[HTML]{009901} 64} & {\color[HTML]{009901} .652} & {\color[HTML]{009901} .305} & {\color[HTML]{009901} .444} \\
\multirow{-3}{*}{\textbf{CGE}~\cite{naeem2021learning}} & {\color[HTML]{963400} 96} & {\color[HTML]{963400} .671} & {\color[HTML]{963400} \textbf{.332}} & {\color[HTML]{963400} .409} \\ \hline
 & {\color[HTML]{3531FF} 32} & {\color[HTML]{3531FF} \textbf{689}} & {\color[HTML]{3531FF} \textbf{.320}} & {\color[HTML]{3531FF} \textbf{.846}} \\
 & {\color[HTML]{009901} 64} & {\color[HTML]{009901} \textbf{.704}} & {\color[HTML]{009901} \textbf{.327}} & {\color[HTML]{009901} \textbf{.860}} \\
\multirow{-3}{*}{\textbf{Our ZSLA}} & {\color[HTML]{963400} 96} & {\color[HTML]{963400} \textbf{.717}} & {\color[HTML]{963400} .329} & {\color[HTML]{963400} \textbf{.867}}
\end{tabular}
\label{tab:CZSL_result1}
\end{table}
\end{center}


\begin{table*}[t]
\caption{The extended version of Table
in Section~\ref{sec:reannotate} of the main manuscript
, where we additionally provide the results produced by ZSLA-Base as shown in the red-shaded row  (cf. Appendix~\ref{appendix:ZSLA-base}) and the modified CZSL algorithms as shown in the purple-shaded rows (cf. Appendix~\ref{appendix:CZSL}).}
\resizebox{\textwidth}{!}{
\begin{tabular}{c|cccc|cccc|cccc|cccc|}
\cline{2-17}
 & \multicolumn{4}{c|}{CADAVAE~\cite{CADAVAE}} & \multicolumn{4}{c|}{TFVAEGAN~\cite{tfVAEGAN}} & \multicolumn{4}{c|}{ALE~\cite{ALE}} & \multicolumn{4}{c|}{ESZSL~\cite{ESZSL}} \\ \cline{2-17} 
 & S & U & H & {\color[HTML]{B234FC} \textbf{GAIN}} & S & U & H & {\color[HTML]{B234FC} \textbf{GAIN}} & S & U & H & {\color[HTML]{B234FC} \textbf{GAIN}} & S & U & H & {\color[HTML]{B234FC} \textbf{GAIN}} \\ \hline
\rowcolor[HTML]{ECF4FF} 
\multicolumn{1}{|c|}{\cellcolor[HTML]{ECF4FF}\begin{tabular}[c]{@{}c@{}}Manual\\ ($N^s$=32 for CUB)\end{tabular}} & 42.9 & 27.3 & 33.4 & - & 45.5 & 31.2 & 37.1 & - & 26.4 & 9.2 & 13.7 & - & 29.8 & 10.8 & 15.9 & - \\
\rowcolor[HTML]{E6FFE6} 
\multicolumn{1}{|c|}{\cellcolor[HTML]{E6FFE6}\begin{tabular}[c]{@{}c@{}}Manual\\ ($N^s$=312 for CUB)\end{tabular}} & {\color[HTML]{3531FF} \textbf{53.5}} & {\color[HTML]{330001} 51.6} & {\color[HTML]{3531FF} \textbf{52.4}} & {\color[HTML]{B234FC} +19.0} & {\color[HTML]{FE0000} \textbf{64.7}} & {\color[HTML]{000000} 52.8} & {\color[HTML]{FE0000} \textbf{58.1}} & {\color[HTML]{B234FC} {\ul \textbf{+21.0}}} & {\color[HTML]{FE0000} \textbf{62.8}} & {\color[HTML]{3531FF} \textbf{23.7}} & {\color[HTML]{000000} 34.4} & {\color[HTML]{B234FC} +20.7} & {\color[HTML]{3531FF} \textbf{63.8}} & {\color[HTML]{000000} 12.6} & {\color[HTML]{000000} 21.0} & {\color[HTML]{B234FC} +5.1} \\
\rowcolor[HTML]{FFF3E4} 
\multicolumn{1}{|c|}{\cellcolor[HTML]{FFF3E4}A-LAGO} & {\color[HTML]{000000} 45.4} & 55.4 & 49.9 & {\color[HTML]{B234FC} +16.5} & {\color[HTML]{330001} 57.4} & {\color[HTML]{3531FF} \textbf{53.0}} & 55.1 & {\color[HTML]{B234FC} +18.0} & 51.8 & {\color[HTML]{3531FF} \textbf{27.2}} & {\color[HTML]{3531FF} \textbf{35.6}} & {\color[HTML]{B234FC} +21.9} & 49.7 & {\color[HTML]{FE0000} \textbf{17.1}} & {\color[HTML]{3531FF} \textbf{25.4}} & {\color[HTML]{B234FC} +9.5} \\
\rowcolor[HTML]{FFF3E4} 
\multicolumn{1}{|c|}{\cellcolor[HTML]{FFF3E4}A-ESZSL} & 41.5 & 48.7 & 44.8 & {\color[HTML]{B234FC} +11.4} & 56.0 & 48.5 & 52.0 & {\color[HTML]{B234FC} +14.9} & 46.1 & 19.0 & 26.9 & {\color[HTML]{B234FC} +13.2} & {\color[HTML]{000000} 61.3} & 9.2 & 16.0 & {\color[HTML]{B234FC} +0.1} \\
\rowcolor[HTML]{FFF3E4} 
\multicolumn{1}{|c|}{\cellcolor[HTML]{FFF3E4}\begin{tabular}[c]{@{}c@{}}Our ZSLA\\ ($N^s$=32,  $N^u$=207 for $\delta$-CUB)\end{tabular}} & {\color[HTML]{333333} 50.3} & {\color[HTML]{3531FF} \textbf{56.4}} & {\color[HTML]{FE0000} \textbf{53.2}} & {\color[HTML]{B234FC} {\ul \textbf{+19.8}}} & {\color[HTML]{3531FF} \textbf{59.0}} & {\color[HTML]{FE0000} \textbf{55.9}} & {\color[HTML]{3531FF} \textbf{57.4}} & {\color[HTML]{B234FC} +20.3} & {\color[HTML]{3531FF} \textbf{52.4}} & {\color[HTML]{FE0000} \textbf{27.5}} & {\color[HTML]{FE0000} \textbf{36.1}} & {\color[HTML]{B234FC} {\ul \textbf{+22.4}}} & {\color[HTML]{FE0000} \textbf{65.1}} & {\color[HTML]{3531FF} \textbf{16.4}} & {\color[HTML]{FE0000} \textbf{26.2}} & {\color[HTML]{B234FC} {\ul \textbf{+10.3}}} \\
\rowcolor[HTML]{F5DFFF} 
\multicolumn{1}{|c|}{\cellcolor[HTML]{F5DFFF}RedWine~\cite{misra2017red}} & 46.8 & {\color[HTML]{FE0000} \textbf{59.6}} & 45.4 & {\color[HTML]{B234FC} +12.0} & 54.0 & 49.7 & 51.8 & {\color[HTML]{B234FC} +14.7} & 39.0 & 23.2 & 29.0 & {\color[HTML]{B234FC} +15.3} & 59.6 & 14.1 & {\color[HTML]{000000} 22.8} & {\color[HTML]{B234FC} +6.9} \\
\rowcolor[HTML]{F5DFFF} 
\multicolumn{1}{|c|}{\cellcolor[HTML]{F5DFFF}LE+~\cite{misra2017red}} & 46.6 & 41.3 & {\color[HTML]{000000} 43.8} & {\color[HTML]{B234FC} +10.4} & 54.6 & 44.3 & {\color[HTML]{000000} 48.9} & {\color[HTML]{B234FC} +11.8} & 41.0 & 22.8 & {\color[HTML]{000000} 29.3} & {\color[HTML]{B234FC} +15.6} & 43.4 & 15.0 & {\color[HTML]{000000} 22.2} & {\color[HTML]{B234FC} +6.3} \\
\rowcolor[HTML]{F5DFFF} 
\multicolumn{1}{|c|}{\cellcolor[HTML]{F5DFFF}AttOps~\cite{nagarajan2018attributes}} & {\color[HTML]{FE0000} \textbf{56.0}} & 28.0 & 37.3 & {\color[HTML]{B234FC} +3.9} & 54.7 & 40.8 & 46.7 & {\color[HTML]{B234FC} +9.6} & 45.3 & 14.3 & 21.7 & {\color[HTML]{B234FC} +8.0} & 53.5 & 5.6 & 10.1 & {\color[HTML]{B234FC} -5.8} \\
\rowcolor[HTML]{F5DFFF} 
\multicolumn{1}{|c|}{\cellcolor[HTML]{F5DFFF}\begin{tabular}[c]{@{}c@{}}CGE~\cite{naeem2021learning}\\ ($N^s$=32,  $N^u$=207)\end{tabular}} & 41.7 & 39.2 & 40.4 & {\color[HTML]{B234FC} +7.0} & 58.5 & 36.3 & 44.8 & {\color[HTML]{B234FC} +7.7} & 40.6 & 17.7 & 24.7 & {\color[HTML]{B234FC} +11.0} & 58.6 & 8.7 & 15.1 & {\color[HTML]{B234FC} -0.8} \\
\rowcolor[HTML]{FFCCC9} 
\multicolumn{1}{|c|}{\cellcolor[HTML]{FFCCC9}\begin{tabular}[c]{@{}c@{}}ZSLA-Base\\ ($N=31$, $N^u=239$)\end{tabular}} & 33.7 & 36.2 & 34.9 & {\color[HTML]{B234FC} +1.5} & 57.4 & 24.0 & 33.9 & {\color[HTML]{B234FC} -3.2} & 19.1 & 13.5 & 15.8 & {\color[HTML]{B234FC} +2.1} & 36.0 & 8.9 & 14.2 & {\color[HTML]{B234FC} -1.7} \\ \hline
\end{tabular}
}

\label{tab:ZSLA-Base-GZSL}
\end{table*}

\subsubsection{Train Base Attribute Detectors Directly}
\label{appendix:ZSLA-base}
\modify{We already show that novel attributes $\mathbf{A}^u$ could be obtained via our ZSLA, which starts by training several seen attributes $\mathbf{A}^s$, then decomposes them into base attribute $\mathbf{B}^c$ and $\mathbf{B}^p$, and finally reassembles $\mathbf{B}^c$ and $\mathbf{B}^p$ as $\mathbf{A}^u$. Here, we conduct experiments to discuss if we can train base attribute detectors (i.e. detector of $\mathbf{B}^c$ and $\mathbf{B}^p$) directly rather than using seen attribute $\mathbf{A}^s$ and the intersection model $\mathbb{I}$.
For example, instead of learning detectors for ``blue crown'', ``red crown'' and the intersection model to get ``crown'', we use the given annotations to train base attribute detector ``crown''. }
\modify{Since images in CUB and $\alpha$-CLEVR are described by attributes, we first turn the ``attribute''-wise annotations in both datasets into ``base attribute''-wise ones (e.g. a bird with ``blue crown'' annotation will be re-labeled as ``is blue'' and ``has crown''). Then we utilize these base attribute annotations and apply the method we proposed in Section~\ref{sec: stage_1} 
to train the corresponding base attribute detectors, which can then be assembled into novel attributes detectors by our union model $\mathbb{U}$. 
}

\begin{table}[h!]
\centering
\caption{Evaluation of synthesized novel/unseen attributes on attribute classification (mAUROC), retrieval (mAP@50), and localization (mLA). $N$ is the number of questions that human annotators have to answer when building CUB dataset. In other words, $N$ is the number of seen attributes~(i.e. $N^s$) for A-ESZSL, A-LAGO and ZSLA; is the number of base attribute for ZSLA-Base.The highest scores are marked in bold \textcolor{red}{red}, while the second-highest ones are marked in bold \textcolor{blue}{blue}.}
\begin{tabular}{c|c|ccc}
 & $N$ & mAUROC & mAP@50 & mLA \\ \hline
A-ESZSL & 32 & .626 & .223 & .756 \\ \hline
A-LAGO & 32 & .600 & .173 & .782 \\ \hline
ZSLA & 32 & {\color[HTML]{FF0000} \textbf{.689}} & {\color[HTML]{FF0000} \textbf{.320}} & {\color[HTML]{FF0000} \textbf{.846}} \\ \hline
ZSLA-Base & 31 & {\color[HTML]{0000FF} \textbf{.639}} & {\color[HTML]{0000FF} \textbf{.243}} & {\color[HTML]{0000FF} \textbf{.833}}
\end{tabular}
\vspace{.5em} 
\label{tab:ZSLA-Base}
\end{table}

\modify{In order to distinguish this approach from our ZSLA, we name it \textbf{ZSLA-Base} and conduct experiments to measure its performance and robustness. For CUB dataset, ZSLA-Base uses 31 base attributes to synthesize 239 attributes, and ZSLA can use 32 seen attributes to generate the other 207 novel ones. The number of seen attribute or seen base attribute annotations could be viewed as the manual-cost to construct the dataset (i.e. imagining the process to annotate attributes: human annotators are requested to answer whether the image has a certain attribute or a base attribute, and the number of questions is just the same as the number of seen attributes or the seen base attributes), we hence conclude that the manual-costs for our ZSLA method and the ZSLA-Base approach are almost on the same level. Table~\ref{tab:ZSLA-Base} evaluates the performance of attribute detectors obtained via ZSLA and ZSLA-Base in terms of mAUROC, mAP@50, and mLA. We can observe that ZSLA-Base outperforms A-ESZSL and A-LAGO, while is inferior to our ZSLA in terms of attribute classification, retrieval, and localization (while our ZSLA and ZSLA-Base are at about the same manual-cost).}

\modify{Regarding the experiment using automatic annotations for learning generalized zero-shot image classification (cf. the experiments in Section~\ref{sec:reannotate} 
of our main manuscript), we simply select 0.5 as the threshold to binarize posterior, since ZSLA-Base synthesizes/generates attribute detectors from the base attributes directly thus the thresholds cannot be decided in the same way as we do for ZSLA (i.e. as described in Section~\ref{sec:reannotate} of the main manuscript where the thresholds are determined by maximizing $TPR-FPR$ over all the seen attributes). 
In Table~\ref{tab:ZSLA-Base-GZSL} we provide the experimental results where the attribute labels are annotated by ZSLA-Base method~(i.e. the red-shaded row). We observe that using ZSLA-Base to automatically annotate the attribute labels is unhelpful for the GZSL algorithms even it has acceptable performance on attribute classification, retrieval, and localization as shown in Table~\ref{tab:ZSLA-Base}. This contradiction is caused by the poor threshold selection due to the lack of reference of seen attributes, which becomes the major problem of ZSLA-Base. 
}

\begin{figure}[h!]
    \centering
    \includegraphics[width=0.47\textwidth]{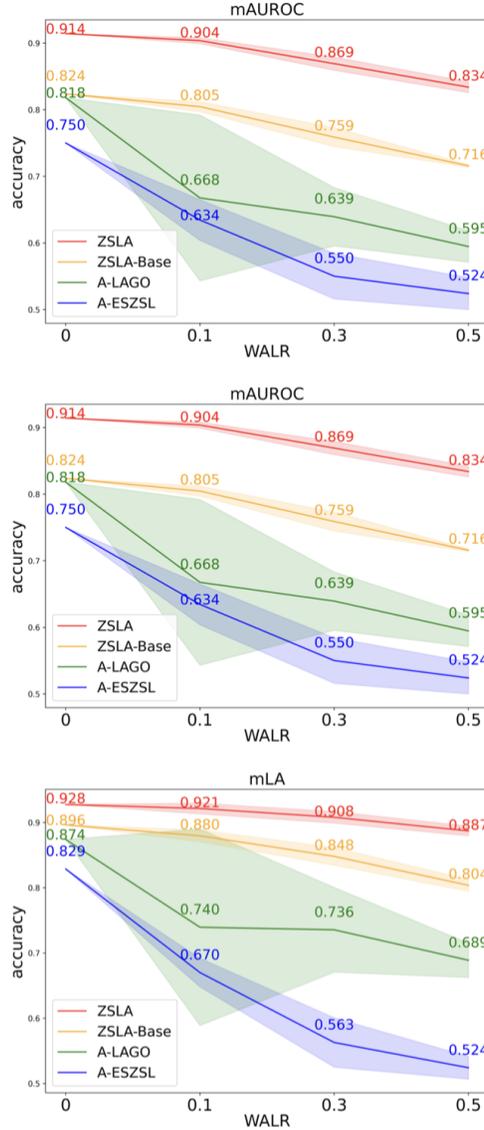}
    \vspace{-1em}
    \caption{\modify{
    Evaluation (in terms of attribute classification, retrieval, and localization, with mAUROC, mAP@50, mLA as metric respectively) on the robustness against noisy attribute labels for various methods which learn to synthesize the novel attributes. The shaded bands around each curve represent the 95\% confidence interval over 5 runs of different noisy label sets.}}
    \label{Fig.ZSLA-Base}
\end{figure}

\begin{figure}[h!]
    \centering
    \includegraphics[width=0.5\textwidth]{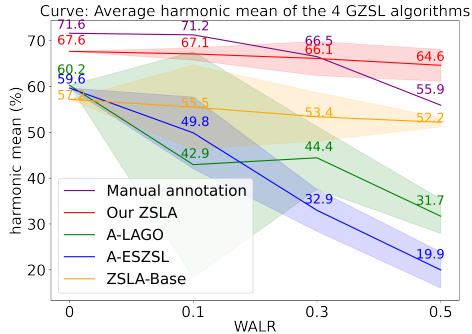}
    \vspace{-0.7em}
    \caption{\modify{Evaluation on the quality of automatic re-annotations produced by different methods, where the performance is based on the average harmonic mean of four GZSL algorithms using the re-annotated attributes (cf. the last paragraph in Section~\ref{sec:robustness} 
    for details).}}
    \label{Fig.MEAN_new_base}
\end{figure}

\modify{Following the experiment setting as described in Section~\ref{sec:robustness} of the main manuscript
, we measure the robustness of ZSLA-Base using the mAUROC, mAP@50, mLA metrics, where the resultant performance curves are shown in Figure~\ref{Fig.ZSLA-Base}. It is clear to observe that: (1) ZSLA and ZSLA-Base surpass other baselines (i.e. A-LAGO and A-ESZSL) in attribute classification, retrieval and localization for all the \textbf{WALRs}, while ZSLA has the best performance; (2) ZSLA and ZSLA-Base are more robust than baselines as indicated by having less performance drop when \textbf{WALR} is increased; (3) both ZSLA and ZSLA-Base has a lower variance than baselines over multiple runs of different noisy label sets. All three statements show that ZSLA-Base is more robust than the other baselines while still having room for improvement compared to our ZSLA.} 

\modify{Furthermore, we utilize ZSLA-Base to synthesize the attribute detectors under various WALR settings, re-annotating  $\alpha$-CLEVR automatically, and use the new $\alpha$-CLEVR to train the four GZSL algorithms, whose result is shown as the orange curve in Figure~\ref{Fig.MEAN_new_base}. It is clear to see that when WALR is set as 0~(i.e. no noisy labeling), ZSLA-Base has the worst performance owing to the improper threshold setting (as described above by two paragraphs ahead); nevertheless, compared to the other baselines (i.e. A-LAGO and A-ESZSL), ZSLA-Base is far more robust as shown in the previous paragraph, hence achieving a better result than the baselines while WALR goes high. To sum up, our proposed ZSLA is no doubt the leading method to automatically annotate instance-level attribute labels over the four approaches~(i.e. ZSLA, ZSLA-Base, A-LAGO, A-ESZSL) owing to its robustness.
}
\subsubsection{Comparison with the baseline of adopting word2vec embedding}
\label{appendix:W2V}

In this section, we demonstrate the comparison in terms of the performance of four GZSL algorithms (i.e. CADAVAE~\cite{CADAVAE}, TFVAEGAN~\cite{tfVAEGAN}, ALE~\cite{ALE}, and ESZSL~\cite{ESZSL}) trained by using either ``attributes as the auxiliary semantics to associate classes'' or ``word2vec as the auxiliary semantics to associate classes'' on the CUB dataset. We can see from the experimental results summarized in Table~\ref{tab:w2vcomp} that, even only leveraging 32 manually annotated attributes to construct the associations across classes (i.e. the blue-shaded row), it already contributes to achieve a better GZSL performance in comparison with using word2vec~\cite{W2V} (i.e. the yellow-shaded row), thus demonstrating the benefits of adopting attributes as the class semantics. However, annotating attributes usually requires expensive cost and that becomes its main burden of applications. To this end, our proposed method of zero-shot learning for attributes directly contributes to alleviate such problem, in which our ZLSA is able to offer additional high quality automatic attribute annotations to construct the zero-shot learning dataset with little cost. For instance in the CUB dataset, given merely 32 seen attributes, we can synthesize another 207 novel attribute detectors for performing attribute annotation, where the promising quality of these additional annotated attributes (acting as the auxiliary semantics to associate classes) is well reflected by the significantly increased GZSL performance (observable in the improvement from the blue-shaded row to the green-shaded row).

\begin{table}[h!]
Comparison with the baseline of adopting word2vec embedding as the class semantic for the GZSL task (where the word2vec embeddings are provided by ~\cite{CADAVAE}) on the CUB dataset. The number in bold red represents the best performance. From the results, we can observe that the class semantics stemming from the attribute annotations produced by our ZSLA (shaded in green) can lead to better performance in comparison to the ones based on word2vec embeddings (shaded in yellow) under four different GZSL algorithms.
\resizebox{\textwidth}{!}{
\begin{tabular}{c|ccc|ccc|ccc|ccc|}
\cline{2-13}
 & \multicolumn{3}{c|}{CADAVAE~\cite{CADAVAE}} & \multicolumn{3}{c|}{TFVAEGAN~\cite{tfVAEGAN}} & \multicolumn{3}{c|}{ALE~\cite{ALE}} & \multicolumn{3}{c|}{ESZSL~\cite{ESZSL}} \\ \cline{2-13} 
 & S & U & H & S & U & H & S & U & H & S & U & H \\ \hline
\rowcolor[HTML]{FFF2CC} 
\multicolumn{1}{|c|}{\cellcolor[HTML]{FFF2CC}word2vec~\cite{CADAVAE}} & {\color[HTML]{FE0000} \textbf{65.5}} & 11.3 & 19.3 & 45.2 & 28.1 & 34.7 & {\color[HTML]{FE0000} \textbf{60.1}} & 3.3 & 6.3 & 63.5 & 1 & 2 \\
\rowcolor[HTML]{C9DAF8} 
\multicolumn{1}{|c|}{\cellcolor[HTML]{C9DAF8}\begin{tabular}[c]{@{}c@{}}Manual\\ ($N^s$=32 for CUB)\end{tabular}} & 42.9 & 27.3 & 33.4 & 45.5 & 31.2 & 37.1 & 26.4 & 9.2 & 13.7 & 29.8 & 10.8 & 15.9 \\
\rowcolor[HTML]{D9EAD3} 
\multicolumn{1}{|c|}{\cellcolor[HTML]{D9EAD3}\begin{tabular}[c]{@{}c@{}}Our ZSLA\\ ($N^s$=32,  $N^u$=207 for $\delta$-CUB)\end{tabular}} & 52.8 & {\color[HTML]{FE0000} \textbf{58.1}} & {\color[HTML]{FE0000} \textbf{55.3}} & {\color[HTML]{FE0000} \textbf{59}} & {\color[HTML]{FE0000} \textbf{55.9}} & {\color[HTML]{FE0000} \textbf{57.4}} & {\color[HTML]{000000} 52.4} & {\color[HTML]{FE0000} \textbf{27.5}} & {\color[HTML]{FE0000} \textbf{36.1}} & {\color[HTML]{FE0000} \textbf{65.1}} & {\color[HTML]{FE0000} \textbf{16.4}} & {\color[HTML]{FE0000} \textbf{26.2}} \\ \hline
\end{tabular}
}
\label{tab:w2vcomp}
\end{table}

\subsection{Supplementary for Limitation} \label{appendix:limitations}

\subsubsection{Dataset}
\modify{
Aside from CUB dataset, Animals with Attributes 2 dataset~\cite{ZSLGBU, lampert2009learning, lampert2013attribute, osherson1991default, kemp2006learning}~(usually abbreviated as AWA2) and Scene Understanding dataset~\cite{SUN} (usually abbreviated as SUN) are also widely used to evaluate the performance of GZSL tasks. Nevertheless, neither AWA2 nor SUN have compound attributes for ZSLA to obtain base attributes by our intersection model (and it is hard to re-annotate the whole datasets to satisfy our experimental settings); therefore, we do not leverage these two datasets in our experiments. In turn, we build an artificial dataset, $\alpha$-CLEVR, in order to further demonstrate the potential of ZSLA and make our experiments convincing. 
}




\modify{In particular, we are expecting that our proposed ZSLA is able to motivate the building of new datasets with the help of ZSLA: Before our work of ZSLA, no matter how the attributes are defined, proposing new datasets requires equally massive costs for manual annotations. Now, our ZSLA provides a new choice: If we can define the attributes of a new dataset in the format that is extendable by the combination of intersection and union operations, ZSLA can help to provide novel attribute annotations in a human-like style during establishing the new datasets and reduce the attribute labeling cost efficiently at the same time.}


\subsubsection{Attribute Synthesis}

\begin{table*}[h!]
\caption{
Extended experimental results of Section~4.2
to study the impact of \textbf{don't care} attributes, where $N^s$ is the number of seen attributes, $N^u$ is the number of unseen attributes and $N^d$ is the number of auxiliary attributes that we view as don't care (where we adopt human annotations to train their detectors)
}
\resizebox{\textwidth}{!}{
\begin{tabular}{c|ccc|ccc|ccc|ccc|}
\cline{2-13}
 & \multicolumn{3}{c|}{CADAVAE~\cite{CADAVAE}} & \multicolumn{3}{c|}{TFVAEGAN~\cite{tfVAEGAN}} & \multicolumn{3}{c|}{ALE~\cite{ALE}} & \multicolumn{3}{c|}{ESZSL~\cite{ESZSL}} \\ \cline{2-13} 
 & S & U & H & S & U & H & S & U & H & S & U & H \\ \hline
\rowcolor[HTML]{FFF3E4} 
\multicolumn{1}{|c|}{\cellcolor[HTML]{FFF3E4}\textbf{$N^s=32, N^u=207$}} & {\color[HTML]{000000} 50.3} & {\color[HTML]{000000} 56.4} & {\color[HTML]{000000} 53.2} & {\color[HTML]{000000} 59.0} & {\color[HTML]{000000} 55.9} & {\color[HTML]{000000} 57.4} & {\color[HTML]{FE0000} \textbf{52.4}} & {\color[HTML]{000000} 27.5} & {\color[HTML]{000000} 36.1} & {\color[HTML]{FE0000} \textbf{65.1}} & {\color[HTML]{000000} 16.4} & {\color[HTML]{000000} 26.2} \\ \hline
\rowcolor[HTML]{ECE4FF} 
\multicolumn{1}{|c|}{\cellcolor[HTML]{ECE4FF}\textbf{$N^s=32, N^u=207, N^d=73$}} & {\color[HTML]{FE0000} \textbf{52.8}} & {\color[HTML]{FE0000} \textbf{58.1}} & {\color[HTML]{FE0000} \textbf{55.3}} & {\color[HTML]{FE0000} \textbf{65.4}} & {\color[HTML]{FE0000} \textbf{59.0}} & {\color[HTML]{FE0000} \textbf{62.0}} & {\color[HTML]{000000} 50.4} & {\color[HTML]{FE0000} \textbf{29.1}} & {\color[HTML]{FE0000} \textbf{36.9}} & {\color[HTML]{000000} 58.0} & {\color[HTML]{FE0000} \textbf{19.9}} & {\color[HTML]{FE0000} \textbf{29.6}} \\ \hline
\end{tabular}
}
\label{tab:deltaCUB_GZSL_don't_care}
\end{table*} 

\modify{
In Section~4.2
, we adopt 32 seen attribute detectors and synthesize 207 novel attribute detectors to re-annotate the CUB dataset as $\delta$-CUB dataset. It is clear to see that $\delta$-CUB has 239 attributes in total, which does not cover the whole 312 attributes in the original CUB. The 73 attributes are excluded owing to the missing common basic concept with the others. For example, the adjective ``hooked'' is only used to describe ``beak''; that is, ``hooked beak'' is the only attribute that contains the base attribute ``hooked''; we thus fail to obtain base attribute ``hooked'' with intersection model. In our setting, we just view these 73 attributes as \textbf{don't care} and observe that we could achieve comparable ~(or even better) GZSL results with 239 attributes only.
However, it is possible that the \textbf{don't care} attributes are important. In this case, we can adopt the first training stage of ZSLA (cf. Section~~\ref{sec: stage_1}) 
to obtain the attribute detectors with the help of human annotations.}

\modify{Here, we conduct an experiment to see what happens if we expand our $\delta$-CUB dataset from 239 attributes to 312 attributes following the above steps (i.e. we adopt human annotations to train the detectors for the \textbf{don't care} attributes) and summarize the results in Table~\ref{tab:deltaCUB_GZSL_don't_care}. From the results, we observe that: with the auxiliary 73 attributes~(i.e. the purple-shaded row of Table~\ref{tab:deltaCUB_GZSL_don't_care}), GZSL algorithms have a subtle improvement compared to ignoring these 73 attributes~(i.e. the orange-shaded row of Table~\ref{tab:deltaCUB_GZSL_don't_care}). The slight advance implies that the extra information for these \textbf{don't care} attributes is useful but might not be so powerful, which explains why we could just omit them in the previous experiments. 
}
\end{document}